\documentclass[letterpaper, 10 pt, conference]{ieeeconf}  

\IEEEoverridecommandlockouts                              

\usepackage{amsmath}
\usepackage{amssymb}
\usepackage{wrapfig}
\usepackage{subfig}
\usepackage{caption}
\usepackage{color}
\usepackage{algorithm}
\usepackage{algorithmic}
\usepackage{graphicx}

\usepackage{pgfplots}

\usepackage{tikz}
\usepackage{tikz-3dplot}

\usepackage{ragged2e}
\usepackage{fp}

\usepackage{ifthen}

\usepackage{tikz}
\usetikzlibrary{shapes.geometric, arrows, calc,automata,positioning,decorations,decorations.text}

\usepackage{multirow}

\usepackage{algorithm}
\usepackage{algorithmic}
\usepackage{graphicx}
\usepackage[english]{babel}
\usepackage{array}
\usepackage[export]{adjustbox}

\graphicspath{{images/}}

\usepackage{xcolor}

\title{\LARGE \bf
Shape Back-Projection In 3D Scenes}

\author{Ashish~Kumar$^{1}$, L. Behera$^{1}$, \textit{Senior Member IEEE},  K. S. Venkatesh$^{1}$\\
{\tt\small {\{\color{red!90!blue}https://github.com/ashishkumar822}\}}
\thanks{$^{1}$Mr. Ashish Kumar, Dr. L. Behera and Dr. K. S. Venkatesh are with the Department of Electrical Engineering, Indian Institute of Technology, Kanpur
        {\tt\small \{krashish,lbehera,venkats\}@iitk.ac.in}}%
}

\begin{document}

\maketitle
\thispagestyle{empty}
\pagestyle{empty}

\begin{justify}

\begin{abstract}
In this work, we propose a novel framework \texttt{shape back-projection} for computationally efficient point cloud processing in a probabilistic manner. The primary component of the technique is \texttt{shape histogram} and a back-projection procedure. The technique measures similarity between 3D surfaces, by analyzing their geometrical properties. It is analogous to color back-projection which measures similarity between images, simply by looking at their color distributions. In the overall process, first, shape histogram of a sample surface (e.g. planar) is computed, which captures the profile of surface normals around a point in form of a probability distribution. Later, the histogram is back-projected onto a test surface and a likelihood score is obtained. The score depicts that how likely a point in the test surface behaves similar to the sample surface, geometrically. Shape back-projection finds its application in binary surface classification, high curvature edge detection in unorganized point cloud, automated point cloud labeling for 3D-CNNs (convolutional neural network) etc. The algorithm can also be used for real-time robotic operations such as autonomous object picking in warehouse automation, ground plane extraction for autonomous vehicles and can be deployed easily on computationally limited platforms (UAVs).

\end{abstract}

\end{justify}

\section{Introduction}
Accurate depth sensing and its efficient processing is crucial for a robotic system to reliably perform tasks such as object pick, place or autonomous navigation. The depth data is acquired by the depth sensors and is commonly known as point clouds. The current depth sensors can provide millions of point cloud data in near real time. However, processing them, in general, requires huge computing power. Hence, driven by the significance of the depth information, in this paper, we focus on exploring local geometrical properties of a point cloud such that multiple tasks in the area of 3D perception can be reasoned by computing only once.
\par
Typically, a point cloud is either organized or unorganized. An organized cloud is represented as a 2D matrix in which each location corresponds to a 3D point similar to a pixel in an image. Such clouds offers straightforward use of the 2D image processing techniques (e.g. edge detection) and facilitates fast nearest neighbor computations. On the other hand, unorganized clouds are simply a collection of 3D points which does not convey any spatial or structural connectivity. These clouds are often represented in form of kD-trees \cite{kdtree} or Octrees \cite{octree} which facilitate efficient (but slower then organized) nearest neighbor search \cite{flann} as well as reduced memory storage. Most common sources of organized point cloud are stereo cameras, Time-of-Flight cameras such as Microsoft Kinect, Intel real sense, IDS-Ensenso etc. Whereas, the LIDAR \cite{velodyne-wp} sensors are a major source of unorganized point clouds. Due to long range and high accuracy, they are preferred in autonomous driving and navigation \cite{ford}, \cite{tesla}, warehouse automation, robotic object manipulation and other industrial applications. 
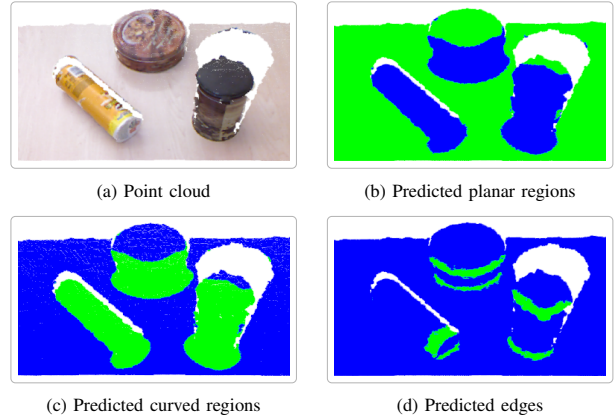
\begin{figure}[t]
\centering

\FPeval{\imheight}{14}
\FPeval{\imwidth}{24}

\FPeval{\xshft}{28}
\FPeval{\yshft}{19}

\FPeval{\lft}{13}
\FPeval{\btm}{9}
\FPeval{\rht}{13}
\FPeval{\tp}{9}

\FPeval{\imscal}{0.95}

\begin{tikzpicture}

\node (n1) [scale= \imscal]{
\includegraphics[width=\imwidth ex,height=\imheight ex, trim={\lft ex, \btm ex, \rht ex, \tp ex}, clip=true]{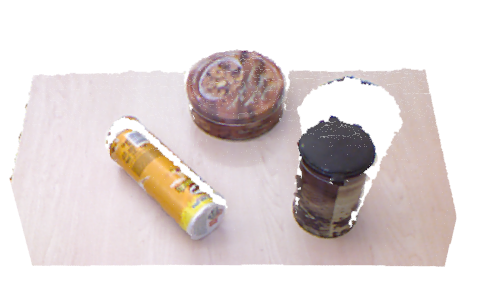}
};

\node (n2) [scale= \imscal,xshift= 1*\xshft ex]{
\includegraphics[width=\imwidth ex,height=\imheight ex, trim={\lft ex, \btm ex, \rht ex, \tp ex}, clip=true]{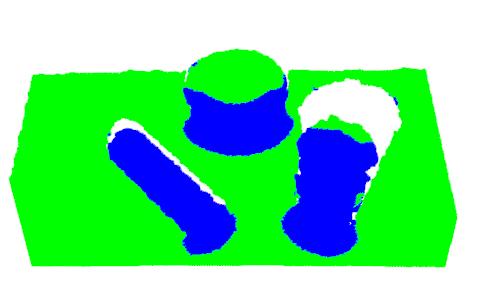}};

\node (n3) [scale= \imscal,xshift= 0*\xshft ex,yshift=-\yshft ex]{
\includegraphics[width=\imwidth ex,height=\imheight ex, trim={\lft ex, \btm ex, \rht ex, \tp ex}, clip=true]{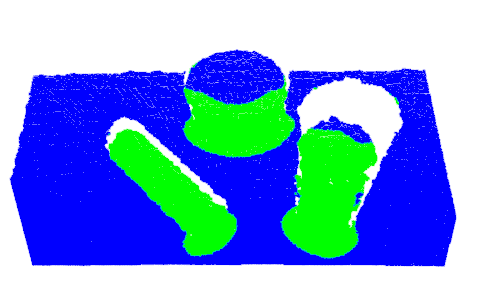}
};

\node (n4) [scale= \imscal,xshift= 1*\xshft ex,yshift=-\yshft ex]{
\includegraphics[width=\imwidth ex,height=\imheight ex, trim={\lft ex, \btm ex, \rht ex, \tp ex}, clip=true]{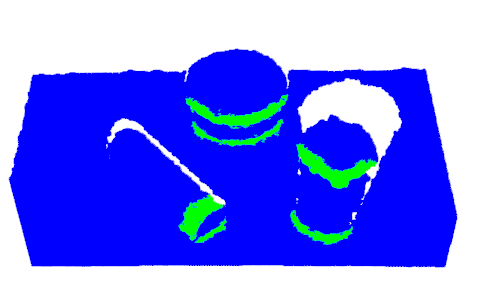}};

\foreach \i in {1,2,...,4}
{
\node (rect\i) at (n\i) [draw=white!70!black, rectangle, rounded corners=0.6mm, minimum width=\imwidth ex , minimum height=\imheight ex, line width=0.1pt, yshift=-0.3ex]{};
}

\node (text1) [below of=n1,yshift=-3ex]{\scriptsize (a) Point cloud};
\node (text2) [below of=n2,yshift=-3ex]{\scriptsize (b) Predicted planar regions};
\node (text3) [below of=n3,yshift=-3ex]{\scriptsize (c) Predicted curved regions};
\node (text4) [below of=n4,yshift=-3ex]{\scriptsize (d) Predicted edges};

\end{tikzpicture}
\caption{Output of the algorithm for tasks such as plane segmentation, curved region segmentation, and edge detection.}
\end{figure}
\par
In order to advance the state-of-art in the above areas, several worldwide robotic challenges have been hosted previously such as Amazon Picking Challenge, 2015 and 2016, Amazon Robotics Challenge, 2017 for warehouse automation, MBZIRC, DARPA humanoid \cite{darpahumanoid} and autonomous driving \cite{darpaautonomous} challenges. Interestingly, all of them have a large technological overlap between them which primarily includes object detection and segmentation of images and point clouds, edge detection (2D or 3D), object pose estimation, 3D model fitting for robot grasping.
\par
The state-of-art (SOA) algorithms for object detection \cite{rcnn}, \cite{fastrcnn}, \cite{fasterrcnn}, \cite{fcn}, \cite{pspnet}, \cite{maskrcnn}, \cite{refinenet} in images are based on Convolutional Neural Network (CNN) architectures \cite{resnet}, \cite{densenet}, \cite{vgg}. The CNNs have also been employed in 3D perception, such as point cloud segmentation \cite{pointcloudseg}, semantic 3D scene completion \cite{semanticscene}, and 3D object segmentation \cite{squeezeseg}. These variants of CNNs are known as 3D-CNNs. The 3D-CNNs require huge amount of labeled point cloud data which comes at a cost of specialized softwares and exhaustive manual efforts. Despite the accuracies, their computational and memory intensive nature limits their scope for real time applications \cite{uav_manipulator}, \cite{uav_manipulator2} on limited computing platforms such as Unmanned Aerial Vehicles (UAVs).
\par
As an alternative, traditional feature matching \cite{fpfh}, \cite{shot} and consensus based model fitting \cite{corres_grp} methods are preferred. The former involve computing of handcrafted features based on local geometrical information of a model point cloud and matching them with the features computed for a target point cloud. The feature matching procedure is a compute intensive task and its performance is severely deteriorated even for minor 3D surface variations between the model and the target. Inconsistent depth data is the primary reason for this which often results in inaccurate feature estimation and matching.
\par
On the other hand, the consensus based methods of RANSAC \cite{ransac}, LMeds \cite{lmeds} are a primary choice to fit primitive shapes (e.g. plane, cylinder or sphere) into point clouds. These methods perform iterative random sampling of the input points and estimates model parameters (plane coefficients, cylinder or sphere radii). However, sampling and the estimation process becomes computationally inefficient when a large number of points are irrelevant to the model to be fit. In case of multiple model instances, the algorithm is iterated exactly equal to number of instances, which in some applications, is not known beforehand. Moreover, their capabilities to only deal with primitive shapes, limit their applicability in real world scenarios, because many objects are often complex shaped, i.e. neither a plane nor a cylinder. 
\par
Furthermore, robust edge detection in the point clouds is quite important for various robotic applications. It can be achieved in the organized point clouds (depth images) simply by using the techniques of edge detection in RGB images. Whereas, unorganized cloud needs special treatment. Such cases are generally occurs when a raw organized point cloud undergoes noise removal preprocessing operations and loses its spatial and structural connectivity (unorganized cloud). Sometimes, the depth sensors (LIDARs) directly provides unorganized point cloud data. Due to the lost spatial relations, edge detection in unorganized point clouds becomes quite challenging and have gained only a little attention \cite{edgedetectunorg}. The approach is based on Eigen value analysis with all results reported only for synthetic data. 
\par
In this paper, we propose a novel probabilistic framework \texttt{Shape Back-projection} which addresses all of the above limitations in a non-iterative manner. 
The most crucial part of the algorithm is shape histograms which encodes the 3D information in such a way that it can be utilized for a number of applications while avoiding complex computations. We experimentally demonstrate that, shape back-projection can be deployed independently for the tasks of point cloud classification, edge detection, especially in the unorganized clouds. Unlike consensus based methods, the proposed algorithm can deal with any number of instances, without manual specification. The algorithm also outperforms a recent algorithm \cite{edgedetectunorg} of edge detection in unorganized point clouds. Moreover, the algorithm can also be effective to facilitate automated labeling of 3D point cloud data, required for 3D-CNNs
\par
In the following sections, first, we lay a preliminary ground on which the whole idea is based (Sec. \ref{preliminaries}). Then, we discuss core of the algorithm (Sec. \ref{algorithm}). At the end, we report comprehensive experimental analysis (Sec.\ref{experiments}) and discuss the primary applications of shape back-projection. 


\section{Preliminaries}
\label{preliminaries}
\subsection{\textbf{Color Histograms}}
\begin{figure}[t]
\centering
\subfloat[]
{
\includegraphics[width=0.18\linewidth,height=0.18\linewidth]{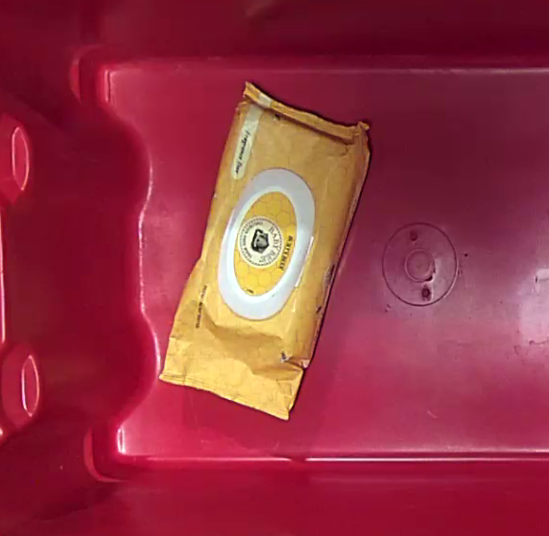}
\label{fig_sample_image}
}
\subfloat[]
{
\begin{tikzpicture}
\begin{axis}[
width=0.60\linewidth,
height=0.40\linewidth,
xlabel={Color intensity},
x label style={at={(axis description cs:0.5,.40)}},
ylabel={P(color$\vert$I)},
y label style={at={(axis description cs:0.45,.5)}},
xmin=0,
xmax=360,
ymin=0,
ymax =.1,
xtick={0,180,360},
scaled y ticks = {base 10:2},
legend cell align={left},
legend pos = north east,
legend style={at={(0.65,0.99)}, anchor=north,font=\tiny, draw=none, row sep=-1ex},
legend image post style={scale=0.2},
label style={font=\fontsize{5}{5}\selectfont},
x tick label style={font=\fontsize{3}{3}\selectfont},
y tick label style={font=\fontsize{3}{3}\selectfont},
scale=0.5
]

\addplot+[fill, mark=none, color= red]
coordinates{(1,8.435790e-04) (2,2.208643e-04) (3,3.834450e-04) (4,3.159587e-04) (5,2.791480e-04) (6,5.951066e-04) (7,3.404992e-04) (8,4.355935e-04) (9,2.454048e-04) (10,6.288498e-04) (11,6.073769e-04) (12,4.355935e-04) (13,5.828364e-04) (14,6.503227e-04) (15,5.920391e-04) (16,7.576873e-04) (17,7.300793e-04) (18,9.141329e-04) (19,9.724165e-04) (20,1.128862e-03) (21,1.082849e-03) (22,1.331321e-03) (23,1.322118e-03) (24,1.613537e-03) (25,1.843604e-03) (26,1.954036e-03) (27,2.245454e-03) (28,2.205576e-03) (29,2.423372e-03) (30,2.690250e-03) (31,3.595180e-03) (32,5.763945e-03) (33,4.917299e-03) (34,5.589094e-03) (35,5.564554e-03) (36,6.401998e-03) (37,6.834524e-03) (38,6.813051e-03) (39,7.316130e-03) (40,6.132052e-03) (41,4.671894e-03) (42,4.859015e-03) (43,4.757785e-03) (44,2.573683e-03) (45,1.865076e-03) (46,1.398807e-03) (47,8.374439e-04) (48,4.662691e-04) (49,4.969447e-04) (50,4.263908e-04) (51,5.337554e-04) (52,1.963238e-04) (53,2.362021e-04) (54,2.484724e-04) (55,3.987828e-04) (56,1.779185e-04) (57,1.288375e-04) (58,5.828364e-05) (59,7.668900e-05) (60,3.067560e-06) (61,5.582959e-04) (62,2.147292e-05) (63,1.840536e-05) (64,1.104322e-04) (65,7.055388e-05) (66,9.816192e-05) (67,6.135120e-05) (68,1.227024e-04) (69,9.816192e-05) (70,3.681072e-05) (71,4.601340e-05) (72,3.987828e-05) (73,6.135120e-05) (74,9.202680e-06) (75,2.454048e-05) (76,6.135120e-05) (77,2.454048e-05) (78,3.987828e-05) (79,3.987828e-05) (80,9.202680e-06) (81,7.055388e-05) (82,2.454048e-05) (83,5.521608e-05) (84,3.067560e-06) (85,1.134997e-04) (86,2.454048e-05) (87,4.601340e-05) (88,3.681072e-05) (89,3.067560e-06) (90,0) (91,4.908096e-05) (92,9.202680e-06) (93,1.533780e-05) (94,1.533780e-05) (95,2.454048e-05) (96,1.533780e-05) (97,2.454048e-05) (98,6.135120e-06) (99,9.202680e-06) (100,3.067560e-06) (101,1.257700e-04) (102,0) (103,2.147292e-05) (104,2.147292e-05) (105,9.202680e-06) (106,1.840536e-05) (107,1.840536e-05) (108,3.067560e-06) (109,3.681072e-05) (110,0) (111,7.668900e-05) (112,6.135120e-06) (113,1.533780e-05) (114,6.135120e-06) (115,0) (116,9.202680e-06) (117,1.840536e-05) (118,6.135120e-06) (119,0) (120,0) (121,1.564456e-04) (122,0) (123,0) (124,3.067560e-06) (125,2.760804e-05) (126,1.533780e-05) (127,1.227024e-05) (128,3.067560e-05) (129,1.533780e-05) (130,0) (131,3.374316e-05) (132,6.135120e-06) (133,1.687158e-04) (134,2.147292e-05) (135,3.067560e-06) (136,5.521608e-05) (137,3.681072e-05) (138,5.214852e-05) (139,1.533780e-05) (140,3.067560e-06) (141,7.055388e-05) (142,3.067560e-05) (143,1.227024e-05) (144,3.067560e-06) (145,8.589168e-05) (146,5.214852e-05) (147,1.840536e-05) (148,1.227024e-05) (149,9.202680e-06) (150,0) (151,1.227024e-04) (152,2.147292e-05) (153,4.294584e-05) (154,2.147292e-05) (155,2.454048e-05) (156,2.147292e-05) (157,2.760804e-05) (158,9.202680e-06) (159,3.374316e-05) (160,3.067560e-06) (161,1.227024e-04) (162,3.067560e-05) (163,1.809860e-04) (164,9.202680e-06) (165,0) (166,4.294584e-05) (167,2.760804e-05) (168,1.227024e-05) (169,1.319051e-04) (170,3.374316e-05) (171,3.374316e-05) (172,3.067560e-05) (173,2.760804e-05) (174,1.227024e-05) (175,1.073646e-04) (176,4.601340e-05) (177,1.227024e-05) (178,9.202680e-06) (179,0) (180,0) (181,4.355935e-04) (182,0) (183,3.067560e-06) (184,3.067560e-06) (185,1.257700e-04) (186,3.681072e-05) (187,9.202680e-06) (188,2.454048e-05) (189,2.546075e-04) (190,9.202680e-05) (191,2.576750e-04) (192,7.975656e-05) (193,2.147292e-05) (194,5.521608e-05) (195,1.319051e-04) (196,1.533780e-05) (197,1.533780e-05) (198,1.441753e-04) (199,2.760804e-05) (200,5.245527e-04) (201,0) (202,3.681072e-05) (203,9.509436e-05) (204,2.177968e-04) (205,4.908096e-05) (206,5.828364e-05) (207,9.816192e-05) (208,1.104322e-04) (209,6.135120e-06) (210,4.816069e-04) (211,0) (212,2.147292e-05) (213,2.147292e-04) (214,8.282412e-05) (215,1.165673e-04) (216,1.073646e-04) (217,7.055388e-05) (218,3.681072e-05) (219,1.840536e-05) (220,1.503104e-04) (221,0) (222,1.134997e-04) (223,3.987828e-05) (224,1.840536e-05) (225,3.895801e-04) (226,1.227024e-05) (227,2.454048e-05) (228,1.656482e-04) (229,3.067560e-06) (230,2.208643e-04) (231,0) (232,2.147292e-05) (233,1.533780e-05) (234,1.533780e-05) (235,9.202680e-06) (236,6.135120e-06) (237,0) (238,0) (239,0) (240,2.914182e-04) (241,0) (242,0) (243,0) (244,6.441876e-05) (245,3.067560e-06) (246,2.147292e-05) (247,6.135120e-06) (248,2.147292e-05) (249,9.202680e-06) (250,1.533780e-05) (251,6.135120e-06) (252,1.595131e-04) (253,6.135120e-06) (254,2.147292e-05) (255,1.196348e-04) (256,0) (257,1.533780e-05) (258,6.135120e-06) (259,1.227024e-05) (260,1.104322e-04) (261,0) (262,9.202680e-06) (263,1.227024e-05) (264,2.454048e-05) (265,0) (266,3.067560e-06) (267,3.067560e-05) (268,0) (269,0) (270,9.509436e-05) (271,0) (272,0) (273,3.067560e-06) (274,0) (275,9.202680e-06) (276,2.147292e-05) (277,0) (278,6.135120e-06) (279,3.067560e-06) (280,2.147292e-05) (281,0) (282,6.135120e-06) (283,9.202680e-06) (284,0) (285,5.828364e-05) (286,3.067560e-06) (287,0) (288,6.135120e-06) (289,3.067560e-06) (290,1.840536e-05) (291,0) (292,9.202680e-06) (293,3.067560e-06) (294,3.067560e-06) (295,0) (296,0) (297,0) (298,0) (299,0) (300,2.147292e-04) (301,0) (302,0) (303,0) (304,0) (305,1.227024e-05) (306,3.067560e-05) (307,9.202680e-06) (308,2.147292e-05) (309,1.227024e-05) (310,9.816192e-05) (311,2.454048e-05) (312,2.454048e-05) (313,2.454048e-05) (314,2.147292e-05) (315,6.135120e-05) (316,2.147292e-05) (317,3.374316e-05) (318,6.135120e-06) (319,2.454048e-05) (320,6.748632e-05) (321,1.227024e-05) (322,3.374316e-05) (323,8.589168e-05) (324,2.607426e-04) (325,1.073646e-04) (326,1.687158e-04) (327,1.963238e-04) (328,3.435667e-04) (329,3.957152e-04) (330,6.533903e-04) (331,8.251736e-04) (332,2.009252e-03) (333,3.678004e-03) (334,3.266951e-03) (335,3.033817e-03) (336,1.023952e-02) (337,1.384083e-02) (338,2.294535e-02) (339,2.500061e-02) (340,2.030418e-02) (341,5.097058e-02) (342,1.232668e-01) (343,7.516749e-02) (344,5.903519e-02) (345,1.571726e-01) (346,1.857408e-01) (347,7.539756e-02) (348,1.996368e-02) (349,1.036835e-02) (350,3.631991e-03) (351,1.797590e-03) (352,9.233355e-04) (353,8.895924e-04) (354,7.331468e-04) (355,5.153501e-04) (356,5.122825e-04) (357,3.865126e-04) (358,2.914182e-04) (359,2.362021e-04) (360,0) } \closedcycle;

\addplot+[fill, mark=none, color= green]
coordinates{(1,7.975656e-05) (2,1.073646e-04) (3,1.687158e-04) (4,2.638102e-04) (5,4.969447e-04) (6,6.902010e-04) (7,6.564578e-04) (8,8.926599e-04) (9,7.239441e-04) (10,9.662814e-04) (11,9.294707e-04) (12,9.785516e-04) (13,8.374439e-04) (14,1.113524e-03) (15,6.503227e-04) (16,6.748632e-04) (17,6.840659e-04) (18,6.196471e-04) (19,7.086063e-04) (20,5.521608e-04) (21,5.276203e-04) (22,4.938772e-04) (23,4.724042e-04) (24,4.632016e-04) (25,3.834450e-04) (26,3.006209e-04) (27,4.355935e-04) (28,3.404992e-04) (29,4.018504e-04) (30,4.263908e-04) (31,3.558370e-04) (32,3.803774e-04) (33,4.539989e-04) (34,4.263908e-04) (35,5.245527e-04) (36,4.417286e-04) (37,4.908096e-04) (38,6.012417e-04) (39,6.073769e-04) (40,6.656605e-04) (41,5.736337e-04) (42,6.503227e-04) (43,7.024712e-04) (44,9.846867e-04) (45,8.159709e-04) (46,7.914305e-04) (47,7.362144e-04) (48,9.049302e-04) (49,1.122727e-03) (50,9.110653e-04) (51,8.558492e-04) (52,7.883629e-04) (53,8.343763e-04) (54,9.969570e-04) (55,8.834573e-04) (56,8.895924e-04) (57,8.926599e-04) (58,9.079977e-04) (59,8.221061e-04) (60,7.730251e-04) (61,7.484846e-04) (62,7.086063e-04) (63,7.116739e-04) (64,6.656605e-04) (65,6.779307e-04) (66,6.871334e-04) (67,6.012417e-04) (68,6.994037e-04) (69,6.963361e-04) (70,6.595254e-04) (71,6.625929e-04) (72,7.055388e-04) (73,6.288498e-04) (74,7.300793e-04) (75,6.687281e-04) (76,8.742546e-04) (77,8.221061e-04) (78,6.994037e-04) (79,6.932685e-04) (80,8.067683e-04) (81,7.270117e-04) (82,1.012295e-03) (83,1.027633e-03) (84,9.233355e-04) (85,9.478760e-04) (86,9.816192e-04) (87,9.724165e-04) (88,9.601463e-04) (89,1.098186e-03) (90,1.205551e-03) (91,1.003092e-03) (92,1.168740e-03) (93,1.199416e-03) (94,9.754841e-04) (95,1.073646e-03) (96,1.282240e-03) (97,1.297578e-03) (98,1.162605e-03) (99,1.208619e-03) (100,1.092051e-03) (101,1.254632e-03) (102,1.181011e-03) (103,1.208619e-03) (104,1.251564e-03) (105,1.230092e-03) (106,1.328253e-03) (107,1.263835e-03) (108,1.616604e-03) (109,1.398807e-03) (110,1.251564e-03) (111,1.490834e-03) (112,1.447888e-03) (113,1.582861e-03) (114,1.849739e-03) (115,1.769982e-03) (116,2.027657e-03) (117,2.067535e-03) (118,2.174900e-03) (119,2.174900e-03) (120,3.279222e-03) (121,2.690250e-03) (122,2.515399e-03) (123,2.677980e-03) (124,2.303738e-03) (125,2.828290e-03) (126,2.619696e-03) (127,2.812952e-03) (128,1.898820e-03) (129,3.089033e-03) (130,3.322167e-03) (131,2.960195e-03) (132,3.003141e-03) (133,3.082898e-03) (134,3.395789e-03) (135,3.408059e-03) (136,3.288424e-03) (137,3.631991e-03) (138,4.018504e-03) (139,4.067584e-03) (140,3.613586e-03) (141,4.319124e-03) (142,4.150409e-03) (143,4.113598e-03) (144,3.929544e-03) (145,3.834450e-03) (146,3.797639e-03) (147,3.653464e-03) (148,3.938747e-03) (149,3.963287e-03) (150,4.092125e-03) (151,4.058382e-03) (152,4.653488e-03) (153,4.966380e-03) (154,4.625880e-03) (155,4.837542e-03) (156,4.972515e-03) (157,5.340622e-03) (158,6.211809e-03) (159,6.086039e-03) (160,5.984809e-03) (161,5.638175e-03) (162,6.067634e-03) (163,5.751675e-03) (164,5.616702e-03) (165,5.524675e-03) (166,5.549216e-03) (167,5.601364e-03) (168,5.901985e-03) (169,5.625905e-03) (170,6.340646e-03) (171,5.058406e-03) (172,5.966404e-03) (173,6.914280e-03) (174,6.156593e-03) (175,5.951066e-03) (176,7.944980e-03) (177,6.230214e-03) (178,7.021645e-03) (179,7.119807e-03) (180,5.101352e-03) (181,7.135144e-03) (182,6.862132e-03) (183,4.567597e-03) (184,7.908170e-03) (185,4.674961e-03) (186,4.592137e-03) (187,5.742472e-03) (188,5.125893e-03) (189,4.359003e-03) (190,6.776240e-03) (191,3.171857e-03) (192,9.196545e-03) (193,5.852904e-03) (194,5.392770e-03) (195,5.267000e-03) (196,5.086014e-03) (197,8.052345e-03) (198,5.254730e-03) (199,6.279295e-03) (200,8.322290e-03) (201,5.236325e-03) (202,8.494073e-03) (203,8.662789e-03) (204,6.843726e-03) (205,1.242362e-02) (206,1.083155e-02) (207,1.257393e-02) (208,1.519976e-02) (209,1.048799e-02) (210,1.362303e-02) (211,1.564149e-02) (212,1.399421e-02) (213,1.602800e-02) (214,1.151255e-02) (215,1.099107e-02) (216,7.849886e-03) (217,7.217969e-03) (218,5.199514e-03) (219,6.997104e-03) (220,5.656581e-03) (221,7.202631e-03) (222,6.622862e-03) (223,8.828438e-03) (224,1.323039e-02) (225,8.343763e-03) (226,1.376721e-02) (227,1.342978e-02) (228,1.592370e-02) (229,1.557400e-02) (230,1.383776e-02) (231,1.730411e-02) (232,1.137144e-02) (233,1.843604e-02) (234,1.540835e-02) (235,1.211073e-02) (236,1.034688e-02) (237,1.369666e-02) (238,6.908145e-03) (239,1.248190e-02) (240,6.113647e-03) (241,1.252485e-02) (242,5.994012e-03) (243,5.122825e-03) (244,2.337481e-03) (245,2.236251e-03) (246,3.770031e-03) (247,3.457140e-03) (248,7.239441e-04) (249,4.573732e-03) (250,1.052173e-03) (251,9.233355e-04) (252,3.337505e-03) (253,3.067560e-04) (254,0) (255,6.632065e-03) 
} \closedcycle;

\addplot+[fill, mark=none, color= blue]
coordinates{(1,0) (2,0) (3,0) (4,0) (5,0) (6,0) (7,3.067560e-06) (8,0) (9,0) (10,0) (11,3.067560e-06) (12,0) (13,3.067560e-06) (14,0) (15,3.067560e-06) (16,0) (17,3.067560e-06) (18,0) (19,0) (20,3.067560e-06) (21,0) (22,0) (23,6.135120e-06) (24,9.202680e-06) (25,3.067560e-06) (26,6.135120e-06) (27,3.067560e-06) (28,1.227024e-05) (29,1.227024e-05) (30,3.067560e-06) (31,9.202680e-06) (32,1.227024e-05) (33,1.227024e-05) (34,1.840536e-05) (35,2.760804e-05) (36,2.760804e-05) (37,3.067560e-05) (38,1.227024e-05) (39,2.147292e-05) (40,9.202680e-06) (41,2.454048e-05) (42,2.147292e-05) (43,6.135120e-06) (44,1.533780e-05) (45,6.748632e-05) (46,3.681072e-05) (47,3.987828e-05) (48,3.067560e-05) (49,3.681072e-05) (50,5.521608e-05) (51,4.908096e-05) (52,6.748632e-05) (53,6.748632e-05) (54,8.589168e-05) (55,1.073646e-04) (56,6.748632e-05) (57,1.134997e-04) (58,1.503104e-04) (59,9.509436e-05) (60,1.257700e-04) (61,1.411078e-04) (62,1.441753e-04) (63,1.564456e-04) (64,1.349726e-04) (65,1.779185e-04) (66,2.147292e-04) (67,1.779185e-04) (68,2.362021e-04) (69,1.809860e-04) (70,2.116616e-04) (71,1.380402e-04) (72,1.257700e-04) (73,1.687158e-04) (74,1.533780e-04) (75,1.963238e-04) (76,1.595131e-04) (77,1.380402e-04) (78,3.957152e-04) (79,1.162605e-03) (80,1.521510e-03) (81,9.969570e-04) (82,1.592064e-03) (83,2.009252e-03) (84,1.889617e-03) (85,3.049155e-03) (86,3.766964e-03) (87,3.515424e-03) (88,2.027657e-03) (89,6.119782e-03) (90,5.224055e-03) (91,3.233208e-03) (92,7.211833e-03) (93,8.119831e-03) (94,8.070750e-03) (95,5.794621e-03) (96,8.144372e-03) (97,6.696483e-03) (98,1.096039e-02) (99,6.616727e-03) (100,6.714889e-03) (101,5.524675e-03) (102,5.395838e-03) (103,7.073793e-03) (104,7.196496e-03) (105,9.730300e-03) (106,8.957275e-03) (107,7.638224e-03) (108,6.877469e-03) (109,6.079904e-03) (110,4.656556e-03) (111,4.714840e-03) (112,5.714864e-03) (113,5.380500e-03) (114,6.653538e-03) (115,4.917299e-03) (116,5.138163e-03) (117,5.444919e-03) (118,4.911163e-03) (119,5.874377e-03) (120,4.168814e-03) (121,3.831382e-03) (122,3.901936e-03) (123,3.914206e-03) (124,5.012393e-03) (125,3.171857e-03) (126,4.420354e-03) (127,4.791529e-03) (128,9.549314e-03) (129,5.610567e-03) (130,5.064541e-03) (131,5.230190e-03) (132,5.466392e-03) (133,4.509313e-03) (134,4.481705e-03) (135,4.119733e-03) (136,4.678029e-03) (137,6.187268e-03) (138,5.374365e-03) (139,5.705661e-03) (140,1.057081e-02) (141,1.131009e-02) (142,1.139905e-02) (143,1.283467e-02) (144,1.509239e-02) (145,1.390218e-02) (146,1.482858e-02) (147,1.772436e-02) (148,1.886243e-02) (149,2.271835e-02) (150,1.608015e-02) (151,1.656176e-02) (152,1.144200e-02) (153,1.024872e-02) (154,9.181207e-03) (155,9.383666e-03) (156,9.073842e-03) (157,9.092248e-03) (158,1.110763e-02) (159,8.518614e-03) (160,9.859138e-03) (161,9.141329e-03) (162,9.291639e-03) (163,7.819210e-03) (164,7.423495e-03) (165,7.954183e-03) (166,7.089131e-03) (167,6.558443e-03) (168,6.733294e-03) (169,6.855996e-03) (170,7.816143e-03) (171,7.702643e-03) (172,7.245577e-03) (173,7.076861e-03) (174,6.377457e-03) (175,7.236374e-03) (176,7.475644e-03) (177,8.037007e-03) (178,7.313063e-03) (179,7.770129e-03) (180,1.016896e-02) (181,1.086530e-02) (182,1.051560e-02) (183,1.095119e-02) (184,9.058505e-03) (185,8.331493e-03) (186,7.791602e-03) (187,7.770129e-03) (188,7.662765e-03) (189,6.699551e-03) (190,6.242484e-03) (191,6.138187e-03) (192,5.990945e-03) (193,5.760878e-03) (194,5.107487e-03) (195,4.978650e-03) (196,5.033866e-03) (197,4.512381e-03) (198,4.309922e-03) (199,4.015436e-03) (200,4.214827e-03) (201,4.276179e-03) (202,3.901936e-03) (203,3.862058e-03) (204,4.178017e-03) (205,3.984760e-03) (206,3.797639e-03) (207,3.613586e-03) (208,3.613586e-03) (209,3.760828e-03) (210,3.785369e-03) (211,3.647329e-03) (212,3.656531e-03) (213,3.766964e-03) (214,3.708680e-03) (215,3.757761e-03) (216,3.659599e-03) (217,3.625856e-03) (218,3.079830e-03) (219,2.543007e-03) (220,2.392697e-03) (221,2.248521e-03) (222,2.131954e-03) (223,1.957103e-03) (224,1.947901e-03) (225,1.635009e-03) (226,1.702496e-03) (227,1.843604e-03) (228,1.601266e-03) (229,1.573658e-03) (230,1.512307e-03) (231,1.460159e-03) (232,1.454023e-03) (233,1.487767e-03) (234,1.408010e-03) (235,1.300645e-03) (236,1.303713e-03) (237,1.165673e-03) (238,1.015362e-03) (239,1.036835e-03) (240,9.264031e-04) (241,9.141329e-04) (242,8.895924e-04) (243,7.791602e-04) (244,7.638224e-04) (245,7.331468e-04) (246,5.276203e-04) (247,5.920391e-04) (248,5.429581e-04) (249,5.092150e-04) (250,6.411200e-04) (251,5.644310e-04) (252,5.981742e-04) (253,5.736337e-04) (254,6.411200e-04) (255,2.297602e-03) } \closedcycle;

\addlegendentry{\tiny Hue}
\addlegendentry{\tiny Sat}
\addlegendentry{\tiny Val}

\end{axis}
\end{tikzpicture}
\label{fig_hist_hsv}
}
\subfloat[]
{
\includegraphics[width=0.18\linewidth,height=0.18\linewidth]{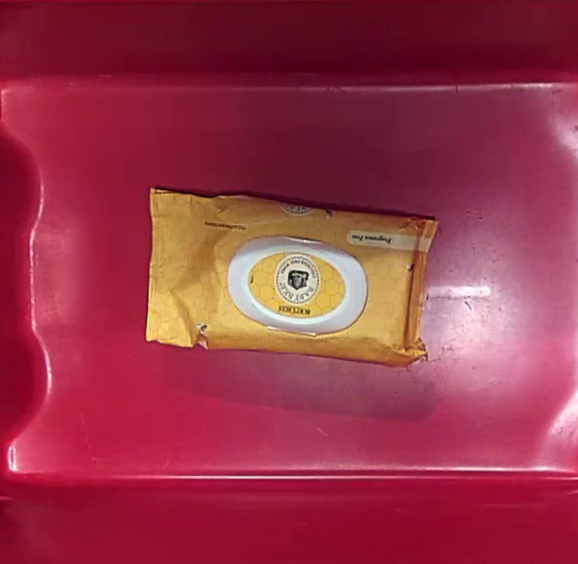}
\label{fig_test_image}
}
\subfloat[]
{
\includegraphics[width=0.18\linewidth,height=0.18\linewidth]{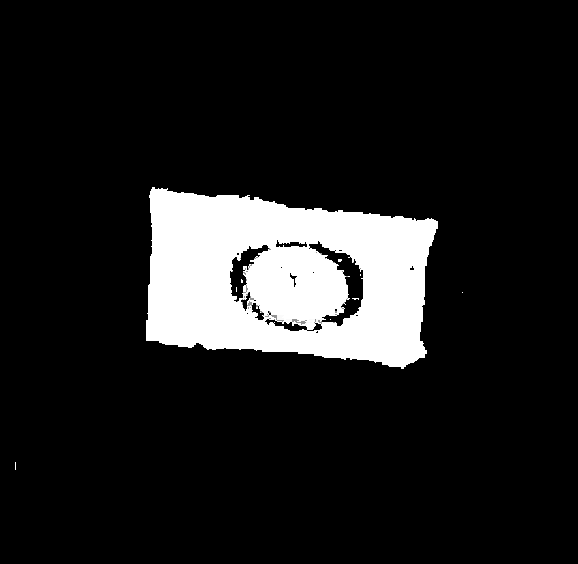}
\label{fig_hist_image}
}
\caption{(a) Sample image, (b) corresponding HSV histogram, (c) test image, and (d) back-projection of the hue histogram of the object pixels in the sample image onto the test image.}
\label{fig_color_hist}
\end{figure}
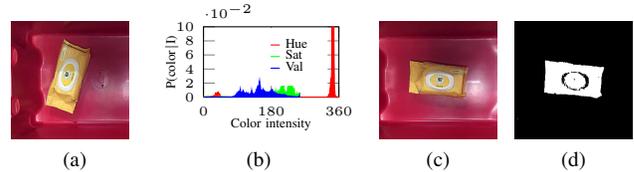
\indent
Color histogram is a discretization of a color space such as (Red, Green, Blue), (Hue, Saturation, Value), etc and represents a frequency distribution over the pixel colors in an image. Each component of a color space is referred as a channel. 
Let $C$ be a $k$-bin color histogram to be computed over a single color channel. First, a $bin$-$id$ (Eq. \ref{eq_bin_id}) for a pixel is obtained and the corresponding bin value is incremented by one. This process is repeated for all or only a desired set of pixels. The obtained frequency distribution ($C$) is then normalized ($C_n$) with the number of pixels which participated in the histogram computation. The normalization step essentially scales all the bin values to sum them up-to one so that it can follow the properties of a valid probability density function. Fig. \ref{fig_hist_hsv} shows single channel normalized histograms of H, S, V components of a sample image (Fig. \ref{fig_sample_image}).
\begin{equation}
bin\text{-}id= \Bigl\lfloor\frac{color}{k}\Bigr\rfloor
\label{eq_bin_id}
\end{equation}
\subsection{\textbf{Color Back-projection}}
\indent
Back-projection \cite{colorbackproj} is a technique to identify test data patterns, behaving almost similar to that of a given distribution. In the context of images, color histograms are back-projected to localize color patterns similar to a given color histogram. In the first step of back-projection, normalized color histogram of the pixels of interest in a sample image is computed (object pixels in Fig. \ref{fig_sample_image}). Later, for each pixel in a test image (Fig. \ref{fig_test_image}), $bin$-$id$ is obtained and assigned a score (Eq. \ref{eq_color_backproj}) equal to the value of bin-id in the color histogram. The score is referred as back-projection likelihood which depicts that how likely a pixel in the test image belongs to the object in the sample image.
\begin{equation}
P(pixel = color~|~C_n) = C_n(bin\text{-}id) 
\label{eq_color_backproj}
\end{equation}
\par
In general, hue component of the HSV color space is preferred for color back-projection because it carries the chrominance information about a pixel-color. However, the selection of color component may vary from application to application. Color back-projection has been explored previously in various applications such as real time object tracking in images using Mean-Shift \cite{meanshiftseg}, Cam-Shift \cite{camshift}.

\section{Shape Back-Projection}
\label{algorithm}
Shape back-projection is inspired by color back-projection and comprises of shape-histogram analogous to a color histogram. By back-projecting the shape histograms onto 3D surfaces, points of a particular interest can be obtained in a probabilistic manner, similar to obtaining pixels of interest in the color back-projection process.
\subsection{\textbf{Shape Histograms}}
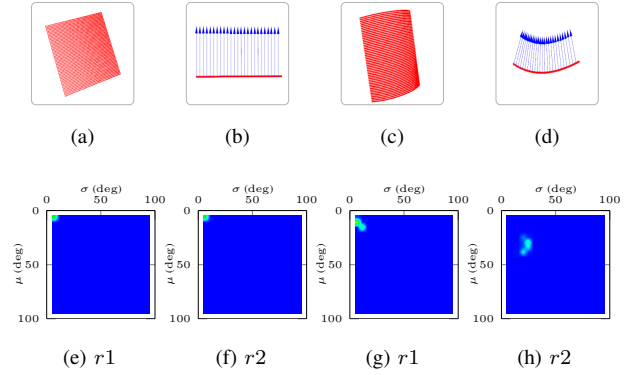
\begin{figure}[t]
\centering

\FPeval{\imwidth}{9}
\FPeval{\imheight}{9}
\FPeval{\shxshft}{0-4.5}
\FPeval{\shscale}{0.8}
\FPeval{\rectoffset}{0.4}
\FPeval{\imxshft}{0.2}

\vspace*{2.7ex}
\subfloat[]
{
\begin{tikzpicture}

\node (im) []{
\includegraphics[width=\imwidth ex,height=\imheight ex]{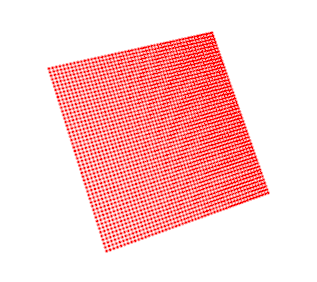}};

\node (rect) at (im) [draw=white!70!black,rounded corners=0.6mm, line width=0.1pt, minimum width=\imwidth ex - \rectoffset ex, minimum height=\imheight ex - \rectoffset ex]{};
\end{tikzpicture}
\label{fig_sample_surface_plane}
}
\hspace{\imxshft ex}
\subfloat[]
{
\begin{tikzpicture}

\node (im) []{
\includegraphics[width=\imwidth ex,height=\imheight ex]{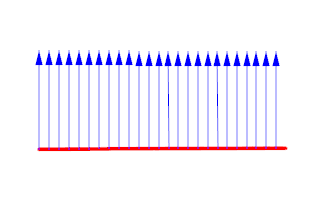}};

\node (rect) at (im) [draw=white!70!black,rounded corners=0.6mm, line width=0.1pt,minimum width=\imwidth ex - \rectoffset ex, minimum height=\imheight ex - \rectoffset ex]{};
\end{tikzpicture}
\label{fig_normals_profile_plane}
}
\hspace{\imxshft ex}
\subfloat[]
{
\begin{tikzpicture}

\node (im) []{
\includegraphics[width=\imwidth ex,height=\imheight ex]{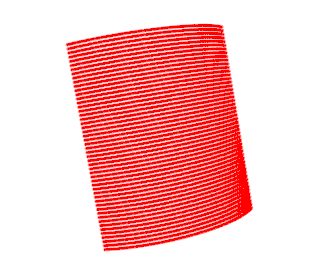}};

\node (rect) at (im) [draw=white!70!black,rounded corners=0.6mm, line width=0.1pt,minimum width=\imwidth ex - \rectoffset ex, minimum height=\imheight ex - \rectoffset ex]{};
\end{tikzpicture}
\label{fig_sample_surface_cylinder}
}
\hspace{\imxshft ex}
\subfloat[]
{
\begin{tikzpicture}

\node (im) []{
\includegraphics[width=\imwidth ex,height=\imheight ex]{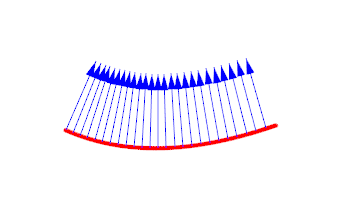}};

\node (rect) at (im) [draw=white!70!black,rounded corners=0.6mm, line width=0.1pt,minimum width=\imwidth ex - \rectoffset ex, minimum height=\imheight ex - \rectoffset ex]{};
\end{tikzpicture}
\label{fig_normals_profile_cylinder}
}
\vspace*{-0.5ex}
\subfloat[$r1$]
{
\begin{tikzpicture}

\node (ex1) [scale=\shscale]{
\begin{tikzpicture}

\begin{axis}[
width=0.60\linewidth,
height=0.60\linewidth,
xlabel={$\sigma$ ($\deg$)},
x label style={at={(axis description cs:0.5,1.60)}},
ylabel={$\mu$ ($\deg$) },
y label style={at={(axis description cs:0.42,.5)}},
xmin=0,
xmax=100,
ymin=0,
ymax =100,
xtick={0,50,100},
ytick={0,50,100},
label style={font=\fontsize{5}{5}\selectfont},
x tick label style={font=\fontsize{3}{3}\selectfont},
y tick label style={font=\fontsize{3}{3}\selectfont},
axis x line*=top,
y dir=reverse,
scale=0.5
]

\draw (axis description cs:0,0) -- (axis description cs:1,0);

\addplot[thick,blue] graphics[xmin=5,ymin=5,xmax=95,ymax=95] {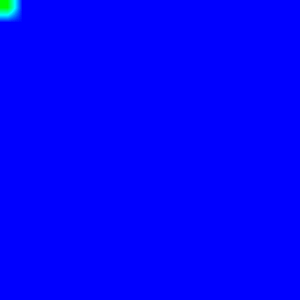};

\end{axis}

\end{tikzpicture}
};
\end{tikzpicture}

\label{fig_sh_plane}

}
\hspace*{\shxshft ex}
\subfloat[$r2$]
{
\begin{tikzpicture}

\node (ex1) [scale=\shscale]{
\begin{tikzpicture}

\begin{axis}[
width=0.60\linewidth,
height=0.60\linewidth,
xlabel={$\sigma$ ($\deg$)},
x label style={at={(axis description cs:0.5,1.60)}},
ylabel={$\mu$ ($\deg$) },
y label style={at={(axis description cs:0.42,.5)}},
xmin=0,
xmax=100,
ymin=0,
ymax =100,
xtick={0,50,100},
ytick={0,50,100},
label style={font=\fontsize{5}{5}\selectfont},
x tick label style={font=\fontsize{3}{3}\selectfont},
y tick label style={font=\fontsize{3}{3}\selectfont},
axis x line*=top,
y dir=reverse,
scale=0.5
]

\draw (axis description cs:0,0) -- (axis description cs:1,0);

\addplot[thick,blue] graphics[xmin=5,ymin=5,xmax=95,ymax=95] {hist_mean_var_planar.png};

\end{axis}

\end{tikzpicture}
};
\end{tikzpicture}

\label{fig_sh_plane1}

}
\hspace*{\shxshft ex}
\subfloat[$r1$]
{
\begin{tikzpicture}

\node (ex1) [scale=\shscale]{
\begin{tikzpicture}

\begin{axis}[
width=0.60\linewidth,
height=0.60\linewidth,
xlabel={$\sigma$ ($\deg$)},
x label style={at={(axis description cs:0.5,1.60)}},
ylabel={$\mu$ ($\deg$) },
y label style={at={(axis description cs:0.42,.5)}},
xmin=0,
xmax=100,
ymin=0,
ymax =100,
xtick={0,50,100},
ytick={0,50,100},
label style={font=\fontsize{5}{5}\selectfont},
x tick label style={font=\fontsize{3}{3}\selectfont},
y tick label style={font=\fontsize{3}{3}\selectfont},
axis x line*=top,
y dir=reverse,
scale=0.5
]

\draw (axis description cs:0,0) -- (axis description cs:1,0);

\addplot[thick,blue] graphics[xmin=5,ymin=5,xmax=95,ymax=95] {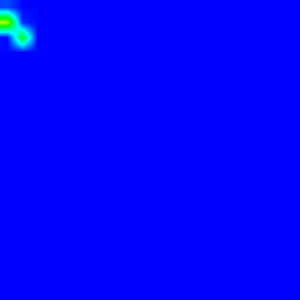};

\end{axis}

\end{tikzpicture}
};
\end{tikzpicture}

\label{fig_sh_cylinder}

}
%
%
%
%
%
%
\hspace*{\shxshft ex}
\subfloat[$r2$]
{
\begin{tikzpicture}

\node (ex1) [scale=\shscale]{
\begin{tikzpicture}

\begin{axis}[
width=0.60\linewidth,
height=0.60\linewidth,
xlabel={$\sigma$ ($\deg$)},
x label style={at={(axis description cs:0.5,1.60)}},
ylabel={$\mu$ ($\deg$)},
y label style={at={(axis description cs:0.42,.5)}},
xmin=0,
xmax=100,
ymin=0,
ymax =100,
xtick={0,50,100},
ytick={0,50,100},
label style={font=\fontsize{5}{5}\selectfont},
x tick label style={font=\fontsize{3}{3}\selectfont},
y tick label style={font=\fontsize{3}{3}\selectfont},
axis x line*=top,
y dir=reverse,
scale=0.5
]

\draw (axis description cs:0,0) -- (axis description cs:1,0);

\addplot[thick,blue] graphics[xmin=5,ymin=5,xmax=95,ymax=95] {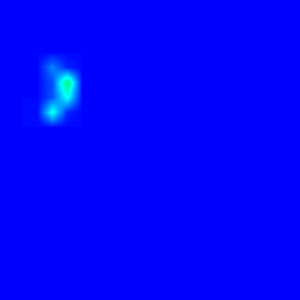};

\end{axis}
\end{tikzpicture}
};
\end{tikzpicture}
\label{fig_sh_cylinder1}
}
%
%
%
%
%
%
\caption{(a), (c) Sample planar and curved surfaces. (b), (d) corresponding profile of surface normals, points (red) and normals(blue). (e), (f) shape histograms of planar surface, and (g), (h) curved surface for different values of neighborhood search radii $r1$ and $r2$, where $r1<r2$ Higher green, higher probability. 
} 
\label{fig_sample_surfaces}
\end{figure}

A shape histogram has been designed by carefully observing the geometrical properties of 3D surfaces such as normals, and curvature. To understand this, consider a planar and a curved surface ($S$) as shown in Fig. \ref{fig_sample_surface_plane} and \ref{fig_sample_surface_cylinder}. Let $p_i,p_j \in S$ be a point and its neighbor respectively and $n_i, n_j$ be their normals. For each ($p_i, p_j$) pair, we define an angle $\alpha_{ij}$ between $n_i$ and $n_j$. An individual $\alpha_{ij}$ does not convey much information, however, the $\alpha_{ij}s$ for all neighbors of $p_i$ governs the behavior of local surface variations around $p_i$. By the visual inspection of Fig.\ref{fig_normals_profile_plane} and \ref{fig_normals_profile_cylinder}, it can be inferred that larger the $\alpha$, more is the surface curved around $p_i$. Based on this fact, we exploit the information contained in the $\alpha_{ij}s$ which we call \textit{Inter Normal Angle Difference} (INAD). We parameterize INAD as a mean and standard deviation ($\mu,\sigma$) value pair of the $\alpha_{ij}s$ for each $p_i$. The INAD pair essentially captures a Gaussian distribution $\mathcal{N}(\mu,\sigma^2)$ of curvature around $p_i$ which is given by Eq.\ref{eq:mean}, \ref{eq:standard_deviation}.
\begin{equation}
\mu_i = \frac{1}{N} \sum_{j=1}^{N} \alpha_{ij}
\label{eq:mean}
\end{equation}
\hfill
\begin{equation}
\sigma_i = \sqrt{\frac{1}{N} \sum_{j=1}^{N} (\alpha_{ij}-\mu_i)^2}
\label{eq:standard_deviation}
\end{equation}             
\par
Further, in order to obtain the shape histogram of a given surface, firstly, INAD for each $p_i \in S$ is computed. Later, a cumulative distribution of these INAD pairs is obtained in a way similar to that of color histogram computation. It is important to note that the INAD is composed of two values: $\mu,\sigma$. Thus, a shape histogram is simply two dimensional and to accommodate this, the Eq.\ref{eq_bin_id} can be rewritten as Eq.\ref{eq_bin_id_sh}, where, $k_\mu, k_\sigma$ are the number of bins in the $\mu$ and $\sigma$ axis
\begin{equation}
bin\text{-}id_{\mu_i}= \Bigl\lfloor\frac{\mu_i}{k_\mu}\Bigr\rfloor,~~~~
bin\text{-}id_{\sigma_i}= \Bigl\lfloor\frac{\sigma_i}{k_\sigma}\Bigr\rfloor
\label{eq_bin_id_sh}
\end{equation}
\par
Next, the obtained cumulative distribution is normalized by dividing it with the maximum value in the distribution.
The normalized cumulative distribution is termed as a \texttt{shape-histogram}. Fig.\ref{fig_sh_plane}-\ref{fig_sh_cylinder1} depicts the shape histogram of a synthetically generated planar and a curved surface for different values of neighborhood search radii $r$.
\par
The INAD computations are entirely based on surface normals. Therefore, noisy normals can severely affect the shape histograms. Most of the noise comes directly from the depth sensors which is aggregated with the approximate solutions errors, introduced by the normal estimation process. Hence, it becomes necessary to deal with the case of noisy normals. Ideally, it would be impossible to obtain 100\% noise free normals. Thus, we employ a noise cancellation procedure based on simple statistical outlier removal technique. Before discussing the noise cancellation procedure, we briefly introduce the normal estimation process below.
%
\subsubsection{\textbf{Normal Estimation}}
Let $p_i$ be any point in $\mathbb{R}^3$ and $N \subset S$ be the set of all neighboring points around $P$ in a spherical region of radii $r$. A covariance matrix ($\mathcal{C}$) in $\mathbb{R}^3$ is computed from a point set $P\cup N$ and decomposed into Eigen vectors. The matrix $\mathcal{C}$ essentially captures the spatial behavior (variance or spread) of neighboring points around $p_i$ and an Eigen vector quantifies both the direction and amount of spread. In general, a surface has minimal spread in a direction normal to it. Therefore, the Eigen vector having lowest Eigen value can be taken as an approximation to the normal at $p_i$. The above methodology is the most simple and fastest approximation of normals and its practical implementations are publicly available in the Point Cloud Library (PCL) \cite{pcl}.
\subsubsection{\textbf{Noise Removal}}
After the normal estimation process, $\alpha_{ij}s$ for each $p_i$ are computed using inverse cosine rule\footnote{$\alpha_{ij}=cos^{-1}\Bigl(\frac{n_i.n_j}{\vert n_i \vert.\vert n_j \vert}\Bigr)$}. At this stage, we assume that some of the $\alpha_{ij}s$ may be noisy. To filter such values, we perform statistical outlier rejection operation on $\alpha_{ij}s$. This step eliminates noisy normals up-to a great extent, leading to consistent normals in the output. The mathematical treatment of the operation is given by Eq. (\ref{eq:mean}, \ref{eq:standard_deviation}, \ref{eq:statistical_outlier}). Here, $c$ is referred as an outlier rate which controls the number of outliers in the output. Decision for a point to be an outlier or inlier is given by Eq. \ref{eq:statistical_outlier}.

\begin{equation}
p_j~\text{is } =
\begin{cases}
Inlier,  & \text{if } \frac{|\alpha_{ij}-\mu_i|}{\sigma_i} \leq c \\
Outlier, & \text{Otherwise}
\end{cases}
\label{eq:statistical_outlier}
\end{equation}
\par
The filtered $\alpha_{ij}s$ are now plugged into Eq. \ref{eq:mean}, \ref{eq:standard_deviation} in order to compute the new values of $\mu_{i}$ and $\sigma_{i}$ which collectively represents INAD at $p_i$. It must be noted that any point $p_j \in N$ marked as an outlier is not included into computation of $\mu_{i}$ and $\sigma_{i}$. Fig. \ref{fig_sh_plane} and \ref{fig_sh_cylinder} shows shape histogram computed for a planar and a curved surface. 
%
%
%
%
%
%
%
%
%
%
%
%
%
%
%
%
\subsection{\textbf{Shape Histogram Back-projection}}
\begin{figure}[t]
\centering
\FPeval{\imwidth}{8}
\FPeval{\imheight}{8}

\FPeval{\imsep}{2.2}

\begin{tikzpicture}

\node (ex1) []
{
\begin{tikzpicture}
\node (n1) []{\includegraphics[width=\imwidth ex, height=\imheight ex]{sample_surface_planar.png}};

\node (n2) [right of = n1, xshift=\imsep ex]{\includegraphics[width=\imwidth ex, height=\imheight ex]{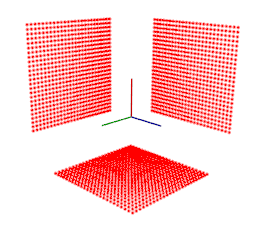}};

\node (n3) [right of = n2,xshift=\imsep ex]{\includegraphics[width=\imwidth ex, height=\imheight ex]{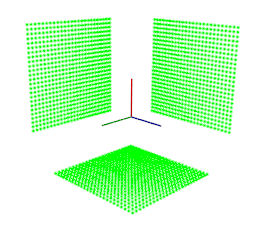}};

\node (rect1) at (n2) [draw=white!70!black, rectangle, rounded corners=0.6mm, minimum width= 25 ex , minimum height=\imheight ex, line width=0.1pt, yshift=0ex]{};

\end{tikzpicture}
};

\node (ex2) [right of = ex1, xshift=21ex]
{
\begin{tikzpicture}

\node (n1) []{\includegraphics[width=\imwidth ex, height=\imheight ex]{sample_surface_planar.png}};

\node (n2) [right of = n1, xshift= \imsep ex]{\includegraphics[width=\imwidth ex, height=\imheight ex]{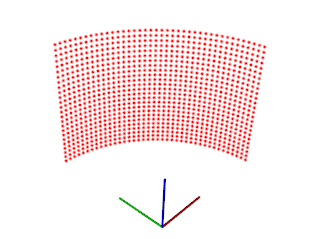}};

\node (n3) [right of = n2,xshift=\imsep ex]{\includegraphics[width=\imwidth ex, height=\imheight ex]{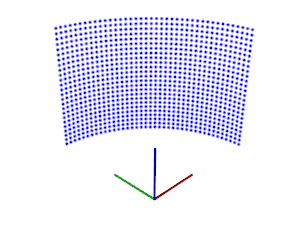}};

\node (rect1) at (n2) [draw=white!70!black, rectangle, rounded corners=0.6mm, minimum width= 25 ex , minimum height=\imheight ex, line width=0.1pt, yshift=0ex]{};

\end{tikzpicture}
};

\node (ex3) [below of =ex1, yshift=-3.5ex]
{
\begin{tikzpicture}

\node (n1) []{\includegraphics[width=\imwidth ex, height=\imheight ex]{sample_surface_cylinder.png}};

\node (n2) [right of = n1, xshift= \imsep ex]{\includegraphics[width=\imwidth ex, height=\imheight ex]{shape_backproj_3_planes.png}};

\node (n3) [right of = n2,xshift=\imsep ex]{\includegraphics[width=\imwidth ex, height=\imheight ex]{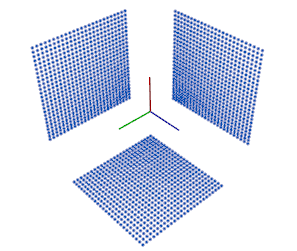}};

\node (rect1) at (n2) [draw=white!70!black, rectangle, rounded corners=0.6mm, minimum width= 25 ex , minimum height=\imheight ex, line width=0.1pt, yshift=0ex]{};

\node (ss) [below of=n1,yshift=1ex]{\scriptsize $S_s$};
\node (st) [below of=n2,yshift=1ex]{\scriptsize $S_t$};
\node (bp) [below of=n3,yshift=1ex]{\scriptsize $P(p \in S_s)$};

\end{tikzpicture}
};

\node (ex4) [right of = ex3, xshift=21ex]
{
\begin{tikzpicture}
\node (n1) []{\includegraphics[width=\imwidth ex, height=\imheight ex]{sample_surface_cylinder.png}};

\node (n2) [right of = n1, xshift=\imsep ex]{\includegraphics[width=\imwidth ex, height=\imheight ex]{shape_backproj_cylinder.png}};

\node (n3) [right of = n2,xshift=\imsep ex]{\includegraphics[width=\imwidth ex, height=\imheight ex]{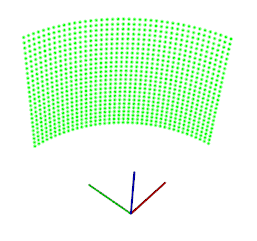}};

\node (rect1) at (n2) [draw=white!70!black, rectangle, rounded corners=0.6mm, minimum width= 25 ex , minimum height=\imheight ex, line width=0.1pt, yshift=0ex]{};

\node (ss) [below of=n1,yshift=1ex]{\scriptsize $S_s$};
\node (st) [below of=n2,yshift=1ex]{\scriptsize $S_t$};
\node (bp) [below of=n3,yshift=1ex]{\scriptsize $P(p \in S_s)$};

\end{tikzpicture}
};



\end{tikzpicture}

\caption{Shape back-projection for various $S_s$-$S_t$ pair}
\label{fig_shape_backproj_examples}
\end{figure}
Shape histogram back-projection is functionally analogous to the color histogram back-projection. In this case, the back-projection likelihood score depicts that: ``how likely a point in a test surface is following geometrical properties similar to that of represented by the shape histogram of a sample surface''. The mathematical expression for such a likelihood is given in Eq.\ref{eq_prob_sh_backproj}, where $S_s$ and $SH$ refers to a sample surface and its shape histogram respectively.
\begin{equation}
P(~p_i \in S_s~|~SH,r) = SH(bin\text{-}id_{\mu_i},~bin\text{-}id_{\sigma_i})
\label{eq_prob_sh_backproj}
\end{equation}
\par
In order to back-project a shape histogram, first the INAD for each point in a test surface $S_t$ is computed (Eq.\ref{eq:mean}, \ref{eq:standard_deviation}). The INAD is then used to obtain $bin\text{-}id_{\mu}$ and $bin\text{-}id_{\sigma}$ which are then plugged into Eq.\ref{eq_prob_sh_backproj} to obtain the likelihood. 
\par
Fig.\ref{fig_shape_backproj_examples} shows examples of the shape back-projection procedure. The column-$1$ represents the case when shape histograms of a planar and a curved surface ($S_s$) are back-projected onto a test surface ($S_t$), having three orthogonal planes. It can be noticed that the likelihood obtained is higher (\raisebox{0.05ex}{\tikz{\node (circle)[fill=green,circle, minimum width=1ex,scale=0.5]{};}}) when both $S_s$ and $S_t$ belong to similar kind of surfaces (planar-planar). Whereas, it is lower (\raisebox{0.05ex}{\tikz{\node (circle)[fill=white!50!blue,circle, minimum width=1ex,scale=0.5]{};}})
when $S_s$ and $S_t$ are of different kinds (curved-planar). Similarly, the column-$2$ represents the case when shape histogram of the planar and curved sample surfaces is back-projected onto a curved test surface $S_t$. 
\par
In the performance of shape back-projection, the neighborhood search radii $r$ is a crucial parameter . The variations in the values of $r$ lead to different utilities of shape back-projection, which we have experimentally demonstrated in the following section.

\section{Experiments}
\label{experiments}
\begin{figure*}[t]
\centering


\FPeval{\imheight}{9}
\FPeval{\imwidth}{15}

\FPeval{\xshft}{16.3}
\FPeval{\yshft}{0-9.5}

\FPeval{\lft}{13}
\FPeval{\btm}{4}
\FPeval{\rht}{13}
\FPeval{\tp}{1}

\FPeval{\imscal}{0.95}

\begin{tikzpicture}

\node (ex1) []{
\begin{tikzpicture}



\node (n1) [scale= \imscal]{
\includegraphics[width=\imwidth ex,height=\imheight ex, trim={\lft ex, \btm ex, \rht ex, \tp ex}, clip=true]{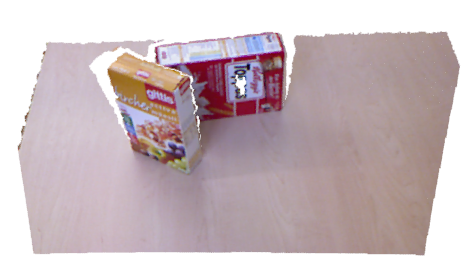}
};

\node (n2) [scale= \imscal,xshift= \xshft ex]{
\includegraphics[width=\imwidth ex,height=\imheight ex, trim={\lft ex, \btm ex, \rht ex, \tp ex}, clip=true]{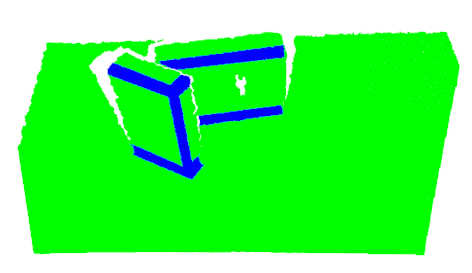}};

\node (n3) [scale= \imscal,xshift= 2*\xshft ex]{
\includegraphics[width=\imwidth ex,height=\imheight ex, trim={\lft ex, \btm ex, \rht ex, \tp ex}, clip=true]{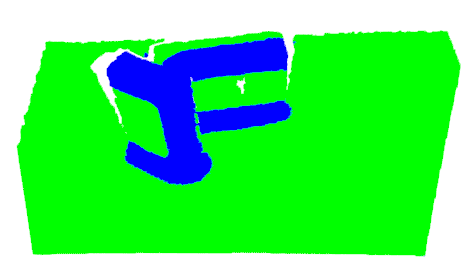}
};

\node (n4) [scale= \imscal,xshift= 3*\xshft ex]{
\includegraphics[width=\imwidth ex,height=\imheight ex, trim={\lft ex, \btm ex, \rht ex, \tp ex}, clip=true]{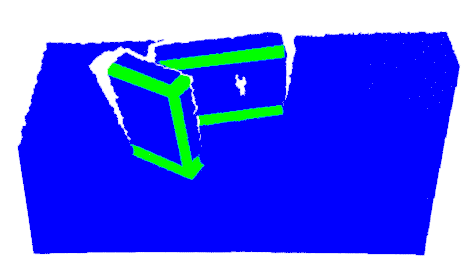}};

\node (n5) [scale= \imscal,xshift= 4*\xshft ex]{
\includegraphics[width=\imwidth ex,height=\imheight ex, trim={\lft ex, \btm ex, \rht ex, \tp ex}, clip=true]{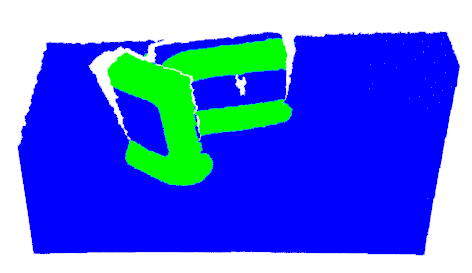}};

\node (n6) [scale= \imscal,xshift= 5*\xshft ex]{
\includegraphics[width=\imwidth ex,height=\imheight ex, trim={\lft ex, \btm ex, \rht ex, \tp ex}, clip=true]{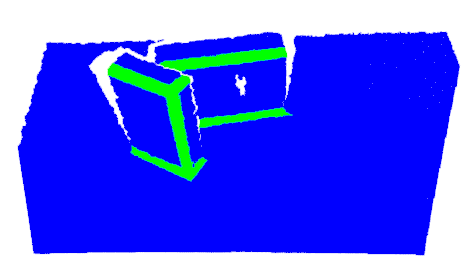}};

\node (n7) [scale= \imscal,xshift= 6*\xshft ex]{
\includegraphics[width=\imwidth ex,height=\imheight ex, trim={\lft ex, \btm ex, \rht ex, \tp ex}, clip=true]{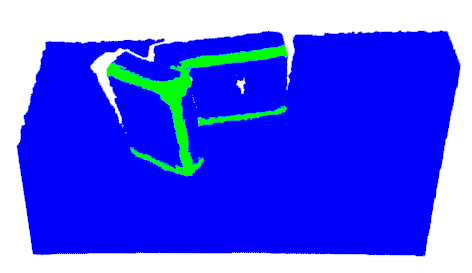}};

\foreach \i in {1,2,...,7}
{
\node (rect\i) at (n\i) [draw=white!70!black, rectangle, rounded corners=0.6mm, minimum width=\imwidth ex , minimum height=\imheight ex, line width=0.1pt, yshift=-0.3ex]{};
}



\end{tikzpicture}
};

\node (ex2) [yshift=\yshft ex]{
\begin{tikzpicture}

\node (n1) [scale= \imscal,]{
\includegraphics[width=\imwidth ex,height=\imheight ex, trim={\lft ex, \btm ex, \rht ex, \tp ex}, clip=true]{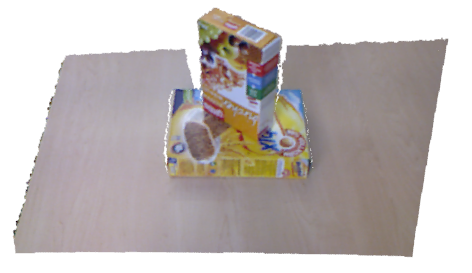}};

\node (n2) [scale= \imscal,xshift= \xshft ex]{
\includegraphics[width=\imwidth ex,height=\imheight ex, trim={\lft ex, \btm ex, \rht ex, \tp ex}, clip=true]{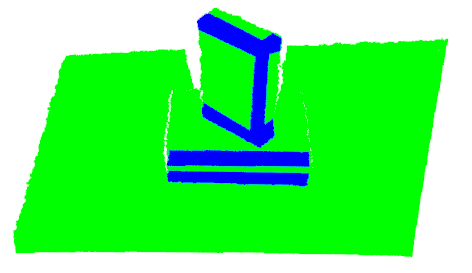}};

\node (n3) [scale= \imscal,xshift= 2*\xshft ex]{
\includegraphics[width=\imwidth ex,height=\imheight ex, trim={\lft ex, \btm ex, \rht ex, \tp ex}, clip=true]{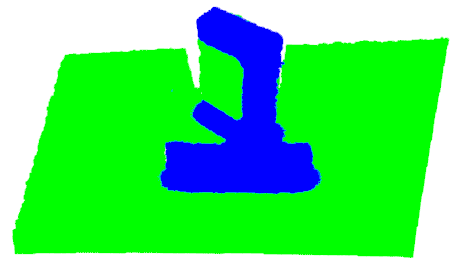}
};

\node (n4) [scale= \imscal,xshift= 3*\xshft ex]{
\includegraphics[width=\imwidth ex,height=\imheight ex, trim={\lft ex, \btm ex, \rht ex, \tp ex}, clip=true]{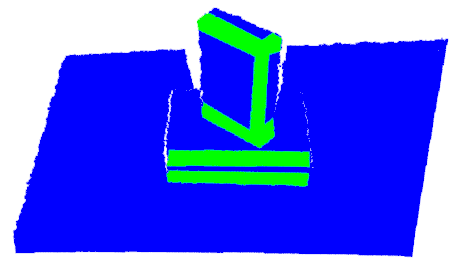}};

\node (n5) [scale= \imscal,xshift= 4*\xshft ex]{
\includegraphics[width=\imwidth ex,height=\imheight ex, trim={\lft ex, \btm ex, \rht ex, \tp ex}, clip=true]{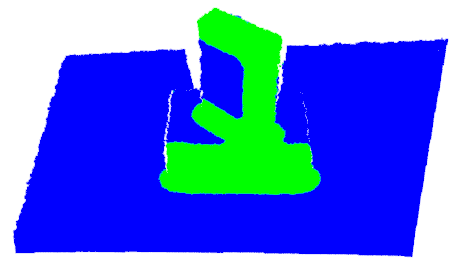}};

\node (n6) [scale= \imscal,xshift= 5*\xshft ex]{
\includegraphics[width=\imwidth ex,height=\imheight ex, trim={\lft ex, \btm ex, \rht ex, \tp ex}, clip=true]{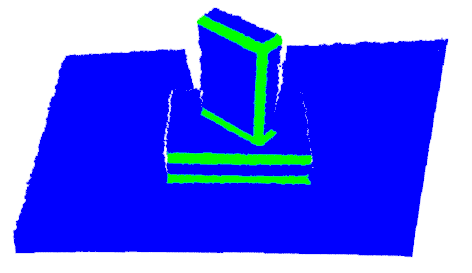}};

\node (n7) [scale= \imscal,xshift= 6*\xshft ex]{
\includegraphics[width=\imwidth ex,height=\imheight ex, trim={\lft ex, \btm ex, \rht ex, \tp ex}, clip=true]{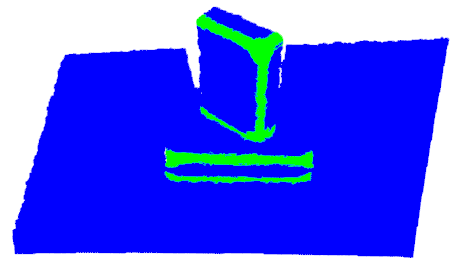}};

\foreach \i in {1,2,...,7}
{
\node (rect\i) at (n\i) [draw=white!70!black, rectangle, rounded corners=0.6mm, minimum width=\imwidth ex , minimum height=\imheight ex, line width=0.1pt, yshift=-0.3ex]{};
}



\end{tikzpicture}
};

\node (ex3) [yshift= 2*\yshft ex]{
\begin{tikzpicture}

\node (n1) [scale= \imscal,]{
\includegraphics[width=\imwidth ex,height=\imheight ex, trim={\lft ex, \btm ex, \rht ex, \tp ex}, clip=true]{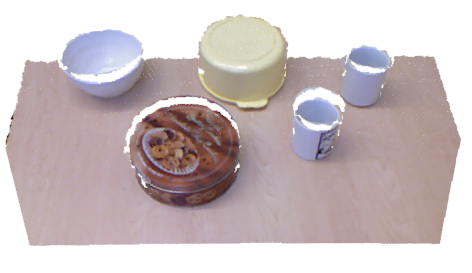}};

\node (n2) [scale= \imscal,xshift= \xshft ex]{
\includegraphics[width=\imwidth ex,height=\imheight ex, trim={\lft ex, \btm ex, \rht ex, \tp ex}, clip=true]{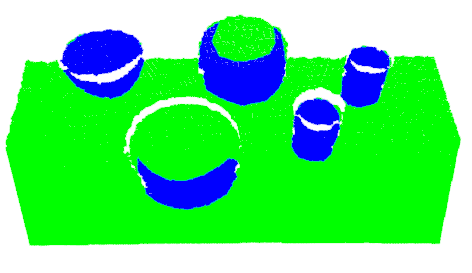}};

\node (n3) [scale= \imscal,xshift= 2*\xshft ex]{
\includegraphics[width=\imwidth ex,height=\imheight ex, trim={\lft ex, \btm ex, \rht ex, \tp ex}, clip=true]{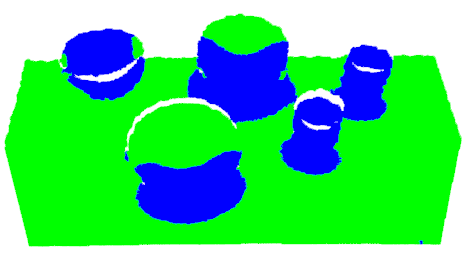}
};

\node (n4) [scale= \imscal,xshift= 3*\xshft ex]{
\includegraphics[width=\imwidth ex,height=\imheight ex, trim={\lft ex, \btm ex, \rht ex, \tp ex}, clip=true]{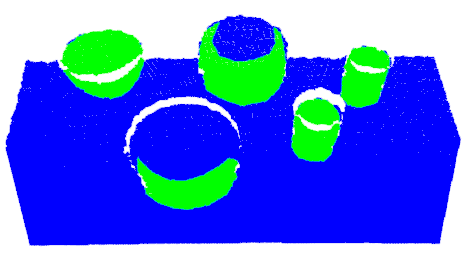}};

\node (n5) [scale= \imscal,xshift= 4*\xshft ex]{
\includegraphics[width=\imwidth ex,height=\imheight ex, trim={\lft ex, \btm ex, \rht ex, \tp ex}, clip=true]{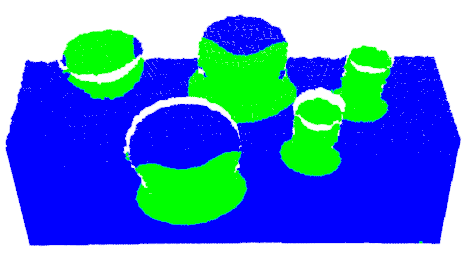}};

\node (n6) [scale= \imscal,xshift= 5*\xshft ex]{
\includegraphics[width=\imwidth ex,height=\imheight ex, trim={\lft ex, \btm ex, \rht ex, \tp ex}, clip=true]{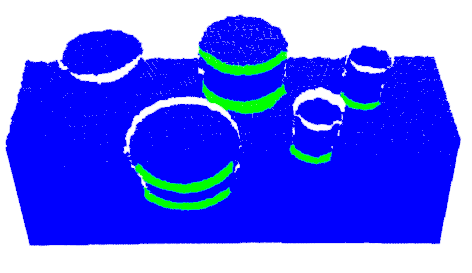}};

\node (n7) [scale= \imscal,xshift= 6*\xshft ex]{
\includegraphics[width=\imwidth ex,height=\imheight ex, trim={\lft ex, \btm ex, \rht ex, \tp ex}, clip=true]{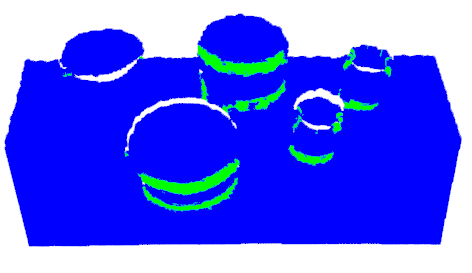}};

\foreach \i in {1,2,...,7}
{
\node (rect\i) at (n\i) [draw=white!70!black, rectangle, rounded corners=0.6mm, minimum width=\imwidth ex , minimum height=\imheight ex, line width=0.1pt, yshift=-0.3ex]{};
}



\end{tikzpicture}
};

\node (ex4) [yshift=3*\yshft ex - 2ex]{
\begin{tikzpicture}

\node (n1) [scale= \imscal,]{
\includegraphics[width=\imwidth ex,height=\imheight ex, trim={\lft ex, \btm ex, \rht ex, \tp ex}, clip=true]{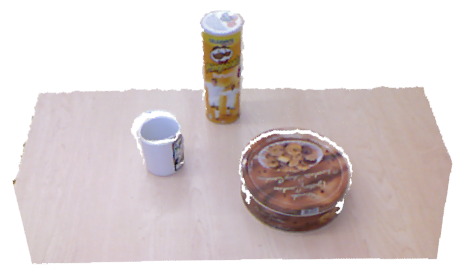}};

\node (n2) [scale= \imscal,xshift= \xshft ex]{
\includegraphics[width=\imwidth ex,height=\imheight ex, trim={\lft ex, \btm ex, \rht ex, \tp ex}, clip=true]{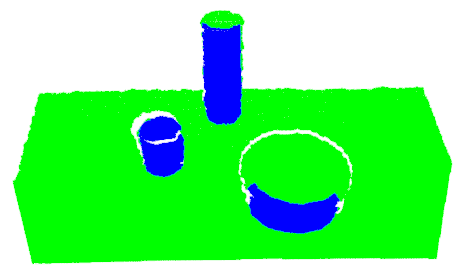}};

\node (n3) [scale= \imscal,xshift= 2*\xshft ex]{
\includegraphics[width=\imwidth ex,height=\imheight ex, trim={\lft ex, \btm ex, \rht ex, \tp ex}, clip=true]{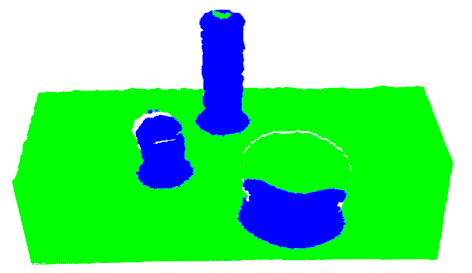}
};

\node (n4) [scale= \imscal,xshift= 3*\xshft ex]{
\includegraphics[width=\imwidth ex,height=\imheight ex, trim={\lft ex, \btm ex, \rht ex, \tp ex}, clip=true]{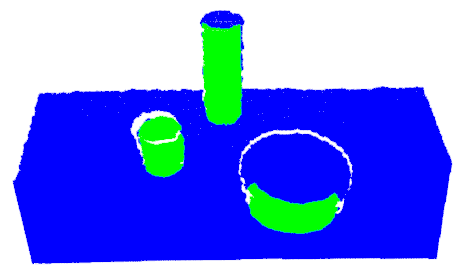}};

\node (n5) [scale= \imscal,xshift= 4*\xshft ex]{
\includegraphics[width=\imwidth ex,height=\imheight ex, trim={\lft ex, \btm ex, \rht ex, \tp ex}, clip=true]{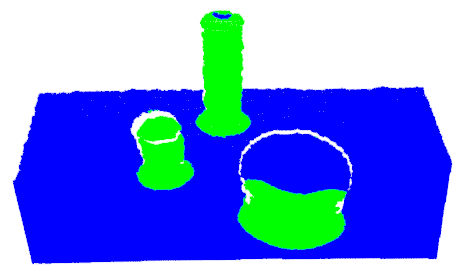}};

\node (n6) [scale= \imscal,xshift= 5*\xshft ex]{
\includegraphics[width=\imwidth ex,height=\imheight ex, trim={\lft ex, \btm ex, \rht ex, \tp ex}, clip=true]{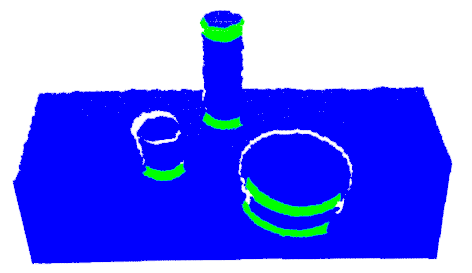}};

\node (n7) [scale= \imscal, xshift= 6*\xshft ex]{
\includegraphics[width=\imwidth ex,height=\imheight ex, trim={\lft ex, \btm ex, \rht ex, \tp ex}, clip=true]{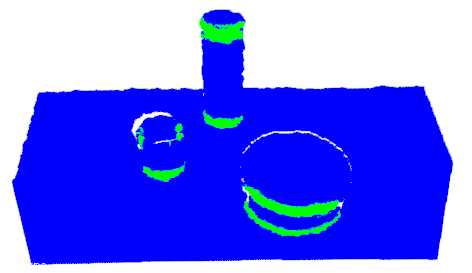}};

\foreach \i in {1,...,7}
{
\node (rect\i) at (n\i) [draw=white!70!black, rectangle, rounded corners=0.6mm, minimum width=\imwidth ex , minimum height=\imheight ex, line width=0.1pt, yshift=-0.3ex]{};
}



\FPeval{\txtyshft}{0-0.4}
\node (text1) [below of=n1, yshift=\txtyshft ex]{\scriptsize \shortstack{Test\\ surfaces}};

\node (text2) [below of=n2, yshift=\txtyshft ex] {\shortstack{\tiny ${P(p \in S_{plane})}$ \\ \scriptsize Ground truth}};

\node (text3) [below of=n3, yshift=\txtyshft ex] {\shortstack{\tiny ${P(p \in S_{plane})}$ \\ \scriptsize Predicted }};

\node (text4) [below of=n4, yshift=\txtyshft ex] {\shortstack{\tiny ${P(p \in S_{curved})}$ \\ \scriptsize Ground truth}};

\node (text5) [below of=n5, yshift=\txtyshft 5ex] {\shortstack{\tiny ${P(p \in S_{curved})}$ \\ \scriptsize Predicted }};

\node (text6) [below of=n6, yshift=\txtyshft ex] {\shortstack{\tiny ${P(p \in S_{edge})}$ \\ \scriptsize Ground truth}};

\node (text7) [below of=n7, yshift=\txtyshft ex] {\shortstack{\tiny ${P(p \in S_{edge})}$ \\ \scriptsize Predicted}};



\end{tikzpicture}
};

\node (colorbar) [xshift= 55.8ex,yshift= -14.5ex]{\includegraphics[width=35ex,height=1.5 ex,rotate=90]{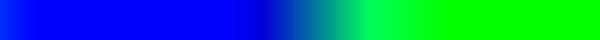}};

\node (colorbar_left) [below of=colorbar,xshift=0ex,yshift= -11.8ex]{\scriptsize{\tiny $\mathbf{0.0}$}};

\node (colorbar_right) [above of=colorbar,xshift= 0ex,yshift= 11.8ex]
{\scriptsize{\tiny $\mathbf{1.0}$}};

\node (colorbar_mid) [ above of=colorbar,xshift= 0ex,yshift= -5ex,rotate=90]
{\scriptsize{\tiny \color{white}$\mathbf{P(p \in Surface \vert SH, r)}$}};



\end{tikzpicture}

\caption{Qualitative results of shape back-projection for binary surface classification and edge detection}
\label{fig_surface_segmentation}
\end{figure*}
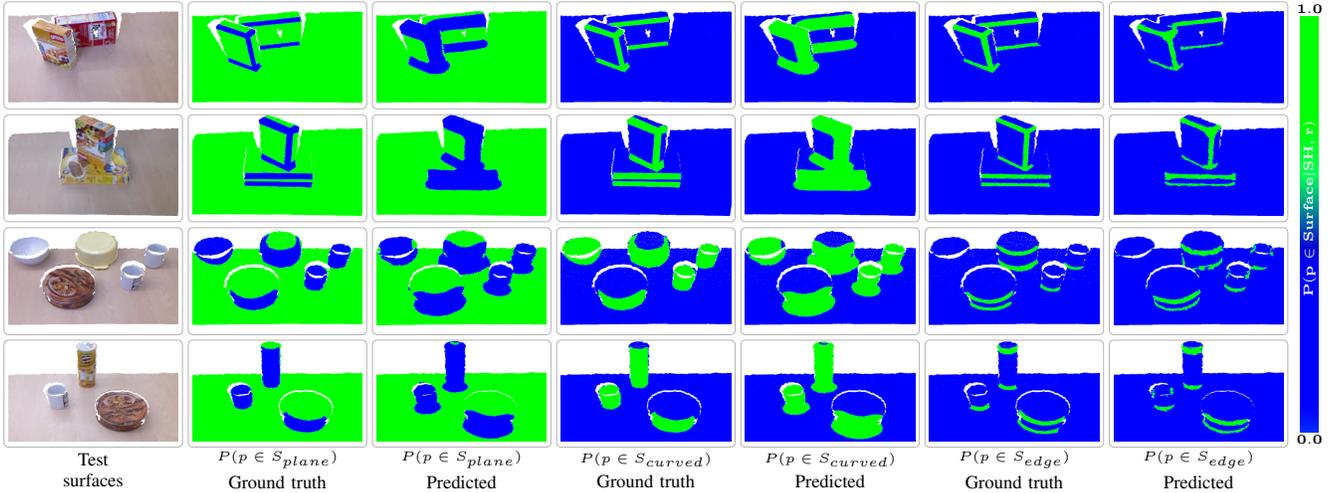
\indent
In this section, we provide quantitative and qualitative results of Shape Back-projection. The experimental evaluation is done by varying the values of parameters such as number of histograms bins ($k$), neighbor search radii ($r$). In order to evaluate the algorithm against noise and demonstrate its real world usefulness, we choose publicly available point cloud dataset \cite{pcl} containing $108$ organized point clouds, acquired by Microsoft Kinect sensor. Each cloud contains a clutter of household objects placed on top of a table. The clouds are deliberately converted into unorganized format. Due to the unavailability of ground truths for the purpose of binary surface classification and edge detection, we manually generate them using CloudCompare \cite{cloudcompare}. All the experiments are performed only on a i7-6850-K CPU and 64GB RAM. It must be noticed that shape-histograms does not represent any feature e.g.\cite{fpfh}, \cite{shot}, therefore we have limited our discussion only to variations in hyper-parameters of shape-histograms and its utilities.
\subsection{\textbf{Varying ``$\mathbf{r}$''}}

The INAD score for a point depends on the number of the spatial neighbors, which in turn is governed by $r$. Fig. \ref{fig_sh_plane}-\ref{fig_sh_cylinder1} shows shape histograms of Fig. \ref{fig_sample_surface_plane} and \ref{fig_sample_surface_cylinder} for varying $r$. It can be seen that as $r$ is increased, the peak in the shape histogram of a planer surface doesn't change whereas it shifts towards higher bins for the case of a curved surface. Thus, we make an important observation that the value of $r$ essentially captures the information about the curvature at a point. We exploit the observation and show that by computing shape histogram of a surface only once, we can use it for different purposes as given below.
\subsubsection{\textbf{Binary Surface Classification}}
Surface segmentation is a most fundamental operation which is required in order to segregate points with similar geometrical properties into different clusters. In this direction, the learning based \cite{squeezeseg}, \cite{pointcloudseg} approaches exists, however, they require tremendous amounts of data and GPUs for their operations. Here, we demonstrate that shape back-projection can achieve high accuracies for binary surface classification merely on a CPU.
\par
The 3D surfaces can be broadly classified into two categories i.e. planar and curved (non-planar). We can say that if a point is more likely to be on a planar surface, it will be less likely to be on a curved surface and vice-versa. Therefore if we have shape histogram of a planar surface, the Eq. \ref{eq_seg_planar_curved} holds.
\begin{equation}
P(p \in S_{plane} \vert SH,r) ~ + ~P(p \in S_{curved} \vert SH, r) ~ = 1
\label{eq_seg_planar_curved}
\end{equation}
\par 
To verify this, we compute shape histogram of a planar surface and back-project onto the real 3D scenes as shown in column-($1$) Fig. \ref{fig_surface_segmentation}. Corresponding color coded likelihoods $P(p \in S_{plane} \vert SH,r)$ and $P(p \in S_{curved} \vert SH, r)$ are shown in column-($2,3$) Fig. \ref{fig_surface_segmentation}. The likelihood score proves the validity of Eq. \ref{eq_seg_planar_curved} qualitatively. The Table \ref{tab_f1_score_seg} depicts the precision, recall, F$1$ and mIoU \cite{voc} scores to assess the binary classification quality. These values are heavily dependent on the quality of the ground truth which we have manually generated. The planar class have high precision and high recall values. Whereas, high recall and relatively low precision for curved surfaces can be accounted by the misclassification of planar points as well as inaccurate ground truths near the high curvature edges. Despite the relatively low precisions for the curved surfaces, the qualitative analysis of binary classification (Fig. \ref{fig_surface_segmentation} column-$2$-$5$) is quite pleasant.
\subsubsection{\textbf{Edge Detection in Unorganized Point Clouds}}
We target this task as a utility of shape back-projection by employing two facts: first, the INAD captures the amount of curvature at a point and second, the edges have high curvature as compared to a plane. Therefore, we reduce $r$ to a smaller value ($\sim6mm$) and then back-project the shape histogram of a planar surface onto the 3D scenes as shown in column-$1$ Fig. \ref{fig_surface_segmentation}. Since, the edges are also curved regions in a cloud, therefore, the Eq. \ref{eq_seg_planar_curved} can be rewritten as given below.
\begin{equation}
P(p \in S_{plane} \vert SH,r) ~ + ~P(p \in S_{edge} \vert SH, r) ~ = 1
\label{eq_edge_planar_curved}
\end{equation}
Table \ref{tab_f1_score_edge} shows the precision, recall and the F$1$-Score for edge detection using shape back-projection and a recent method \cite{edgedetectunorg}. It can be inferred that the recent method performs well for synthetic data which they have reported in their paper, however, performs quite inferior on real data. This is where, the shape back-projection marks its importance to reliably deal with real data. Fig. \ref{fig_surface_segmentation} column-($6$-$7$) shows the color coded edge likelihood $P(p \in S_{edge} \vert SH,r)$ corresponding to the clouds in column-($1$). It can be seen that the predicted edges appears quite similar to the ground truth. 
%
%
%
%
%
%
%
%
%
%
\subsection{\textbf{Number of Bins ($k_\mu,k_\sigma$) and its Effect on Noise}}
\begin{figure}[t]
\centering
\begin{tikzpicture}

\node (n1) []{\includegraphics[scale=0.10,trim={1ex 6ex 1ex 3ex}, clip=true]{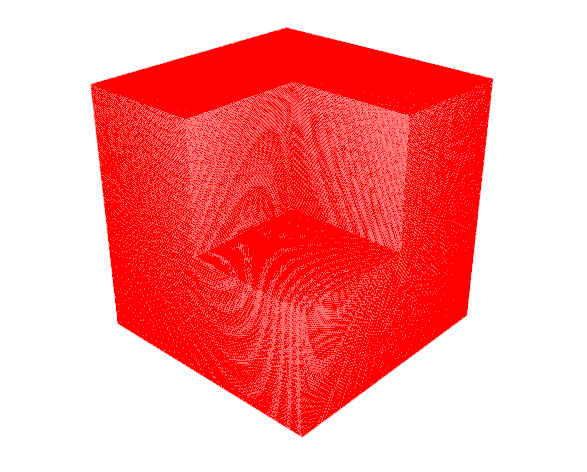}};

\node (n2) [right of =n1, xshift= 10ex]{\includegraphics[scale=0.10,trim={1ex 6ex 1ex 3ex}, clip=true]{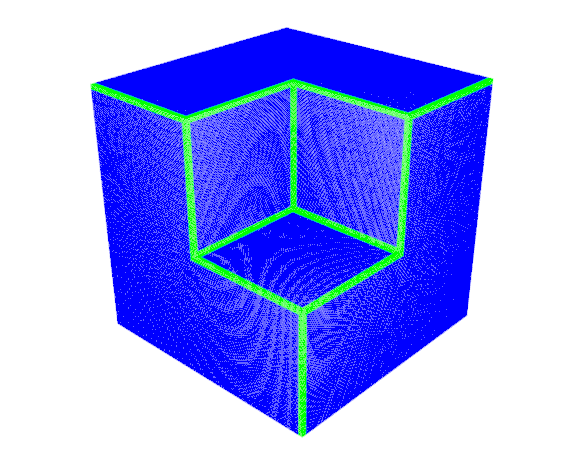}};

\node (n3) [right of =n2, xshift= 10ex]{\includegraphics[scale=0.10,trim={1ex 6ex 1ex 3ex}, clip=true]{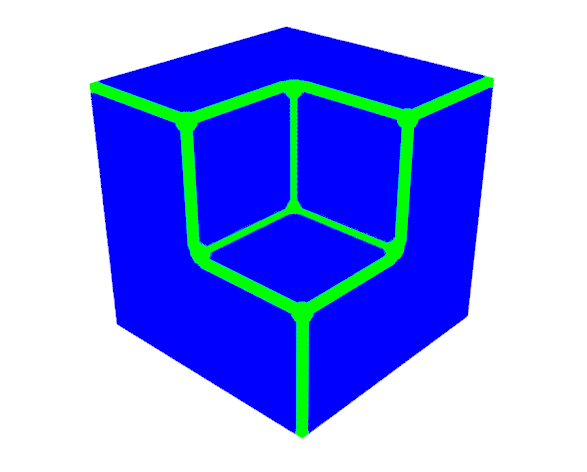}};

\node (text1) [below of=n1]{\scriptsize (a) Synthetic cloud};
\node (text2) [below of=n2]{\scriptsize (b) Ground truth edges};
\node (text3) [below of=n3]{\scriptsize (c) Predicted edges};

\foreach \i in {1,2,3}
{
\node (rect\i) at (n\i) [draw=white!70!black, rectangle, rounded corners=0.6mm, minimum width=10 ex , minimum height=10 ex, line width=0.1pt, yshift=0ex]{};
}

\end{tikzpicture}

\caption{Edge detection in synthetic unorganized point cloud}
\label{fig_synthetic_edge}
\end{figure}
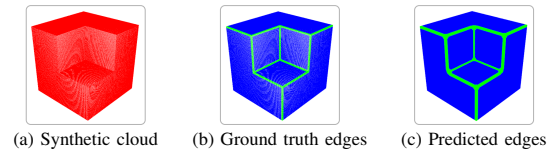

To study the effect of noise, we don't rely on adding noise in the synthetic point clouds as performed in \cite{edgedetectunorg}, instead the chosen point clouds fulfills the purpose, as they are full of random noise and inconsistence depth measurements.
Table \ref{tab_f1_score_edge} depicts the precision, recall and F$1$ measure for high curvature edge detection in the real point clouds data well as a synthetic cloud (Fig. \ref{fig_synthetic_edge}). It is noticeable that the best F$1$ (in \textcolor{blue}{blue}) on real data is obtained with higher number of bins while for the synthetic point cloud, it is achieved with lower number of bins. It happens because noisy point cloud contains local surface variations which can not be represented for small values of $k$. On the other hand, high values of $k$ accurately encodes such information and does not let the noise hamper the underlying geometrical information. Hence, the shape histograms inherently deals with surface noise which is desirable for many practical applications.
\subsection{\textbf{Timing Analysis and Point Cloud Density}}
The INAD score heavily relies on number of nearest neighbors ($k$-NNs) which depends on the cloud density. In general, the real point cloud data is noisy and posses non-uniform density. Therefore, rather changing the radii, timing analysis is performed by explicitly varying the number of nearest neighbors in a synthetic point cloud (Fig. \ref{fig_synthetic_edge}). Table \ref{tab_timing} shows the INAD computing performance for a single point with different number of nearest neighbors. On a scale, $10$ and $500$ neighbors roughly corresponds to a radii of $0.005$m and $0.03$m on surface of resolution $0.001$m. The method is quite fast even on single thread.
This enables shape back-projection to be deployed on computationally limited platforms. The speeds can be increased several times by taking advantage of multi-threading and GPUs, if available.
\subsection{\textbf{Rotation and Translation Invariancy}}
Shape back-projection exploits local geometrical properties. In this process, the INAD score for a point is computed using surface normals which are in turn computed locally and invariant to 3D affine transformation. Therefore, the INAD score and the shape histograms remain rotational and translational invariant.

\subsection{\textbf{A Promising Alternative to Model Consensus}} Consider the point cloud in the Fig.\ref{fig_surface_segmentation} row-$3$. In order to extract all the cylindrical items from the cloud, model consensus needs to be deployed iteratively and number of cylinders (instances) must be known beforehand. On the other hand, binary surface classification using shape back-projection can classify all instances without requiring to have the total number of instances apriori and can also deal with variety of shapes, unlike model consensus which are primitive only. Table \ref{tab_ransac_shapebackproj} shows the precision, recall and F$1$-score of shape back-projection and RANSAC to extract all cylinders in the the the cloud-Id $33$ (Fig.\ref{fig_surface_segmentation}, row-$3$).
%
%
\begin{table*}[t]
\vspace*{0.9ex}

\centering
\caption{\footnotesize Performance analysis of binary surface classification}
\begin{tabular}{|>{\scriptsize }c|>{\scriptsize }c|>{\scriptsize }c|>{\scriptsize }c|>{\scriptsize }c|>{\scriptsize }c|>{\scriptsize }c|>{\scriptsize }c|>{\scriptsize }c|>{\scriptsize }c|>{\scriptsize }c|>{\scriptsize }c|>{\scriptsize }c|>{\scriptsize }c|>{\scriptsize }c|}
\hline
\multirow{4}{*}{Scene-Id} & \multicolumn{14}{c|}{$k_\mu \times k_\sigma$}  \\ \cline{2-15}

 & \multicolumn{7}{c|}{$10 \times 10$} & \multicolumn{7}{c|}{$20 \times 20$}   \\ \cline{2-15}


& \multicolumn{3}{c|}{Planar} & \multicolumn{3}{c|}{Curved} & \multirow{2}{*}{mIoU} & \multicolumn{3}{c|}{Planar} & \multicolumn{3}{c|}{Curved} & \multirow{2}{*}{mIoU} \\ \cline{2-7} \cline{9-14}

& \multicolumn{1}{c|}{Precision} & \multicolumn{1}{c|}{Recall} & \multicolumn{1}{c|}{F1} & \multicolumn{1}{c|}{Precision} & \multicolumn{1}{c|}{Recall} & \multicolumn{1}{c|}{F1} &  &
\multicolumn{1}{c|}{Precision} & \multicolumn{1}{c|}{Recall} & \multicolumn{1}{c|}{F1} &  \multicolumn{1}{c|}{Precision} & \multicolumn{1}{c|}{Recall} & \multicolumn{1}{c|}{F1} & \\ \hline

 $4$ & $0.99$ & $0.95$ & $0.97$ & $0.50$ & $0.98$ & $0.66$ & $0.95$ & $0.99$ & $0.93$ & $0.96$ & $0.38$ & $0.99$ & $0.55$ & $0.93$ \\  
$20$ & $1.00$ & $0.95$ & $0.97$ & $0.63$ & $1.00$ & $0.67$ & $0.95$ & $0.92$ & $0.96$ & $0.92$ & $0.51$ & $1.00$ & $0.68$ & $0.92$\\  
$33$ & $0.97$ & $0.95$ & $0.96$ & $0.74$ & $0.82$ & $0.78$ & $0.92$ & $0.99$ & $0.91$ & $0.95$ & $0.66$ & $0.97$ & $0.79$ & $0.91$ \\  
$44$ & $0.98$ & $0.96$ & $0.97$ & $0.71$ & $0.89$ & $0.79$ & $0.95$ & $0.99$ & $0.93$ & $0.96$ & $0.61$ & $0.99$ & $0.76$ & $0.93$\\ \hline 

\end{tabular}
\label{tab_f1_score_seg}
\end{table*}
\begin{table}[t]

\vspace*{-0.8ex}

\centering
\caption{\footnotesize Performance analysis of high curvature edge detection, S.B. referres to Shape Back-projection }
\begin{tabular}{|>{\scriptsize }c|>{\scriptsize }c|>{\scriptsize }c|>{\scriptsize }c|>{\scriptsize }c|>{\scriptsize }c|>{\scriptsize }c|}
\hline
Scene-Id & Method & $k_\mu$$\times$$k_\sigma$ & Precision & Recall & F$1$ & Time (S) \\ \hline

\multirow{3}{*}{$4$} & S.B. &  $10\times10$ & \textcolor{blue}{$0.98$} & $0.51$ & $0.67$ & $2.2$ \\ 
                     & S.B. &  $20\times20$ & $0.94$ & $0.72$ & \textcolor{blue}{$0.82$} & $2.2$ \\ 
   & \cite{edgedetectunorg} &  - & $0.05$ & $1.00$ & $0.09$ & $2.7$ \\ \hline

\multirow{3}{*}{$20$} & S.B. &  $10\times10$ & \textcolor{blue}{$0.95$} & $0.63$ & $0.75$ & $2.0$ \\ 
                     & S.B. &  $20\times20$ & $0.90$ & $0.83$ & \textcolor{blue}{$0.86$} & $2.0$\\ 
   & \cite{edgedetectunorg} &  - & $0.05$ & $1.00$ & $0.11$ & $2.5$ \\ \hline

\multirow{3}{*}{$33$} & S.B. &  $10\times10$ & \textcolor{blue}{$0.87$} & $0.48$ & $0.62$ & $1.5$ \\ 
                     & S.B. &  $20\times20$ & $0.78$ & $0.75$ & \textcolor{blue}{$0.77$} & $1.5$ \\ 
   & \cite{edgedetectunorg} &  - & $0.05$ & $1.00$ & $0.09$ & $2.1$ \\ \hline

\multirow{3}{*}{$44$} & S.B. &  $10\times10$ & \textcolor{blue}{$0.94$} & $0.62$ & $0.75$ & $1.4$ \\ 
                     & S.B. &  $20\times20$ & $0.84$ & $0.82$ & \textcolor{blue}{$0.83$} & $1.4$ \\ 
   & \cite{edgedetectunorg} &  - & $0.04$ & $1.00$ & $0.08$ & $2.0$ \\ \hline

\multirow{3}{*}{Fig.\ref{fig_synthetic_edge}a} & S.B. &  $10\times10$ & \textcolor{blue}{$0.98$} & $0.98$ & \textcolor{blue}{$0.98$} & $1.5$ \\ 
                     & S.B. &  $20\times20$ & $0.42$ & $0.96$ & $0.59$ & $1.5$  \\ 
   & \cite{edgedetectunorg} &  - & $0.95$ & $0.98$ & $0.97$ & $1.6$ \\ \hline

\end{tabular}
\label{tab_f1_score_edge}
\end{table}
\begin{table}[t]
\centering
\caption{\footnotesize Timing analysis of INAD computations per point}
\label{tab_timing}

\begin{tabular}{|c|c|c|c|c|c|c|}
\hline
$k$-NNs & 10 & 100 & 200 & 300 & 400 & 500 \\ \hline
Time ($\mu$S) & $4.2$ &$23.1$ & $45.5$ &  $67.6$ & $87.5$ & $100.7$ \\ \hline
\end{tabular}

\end{table}
\begin{table}[t]

\vspace*{-0.8ex}

\centering
\caption{\footnotesize Performance analysis for multi-instance cylinder extraction}

\label{tab_ransac_shapebackproj}

\begin{tabular}{|>{\scriptsize }c|>{\scriptsize }c|>{\scriptsize }c|>{\scriptsize }c|}
\hline
Method & Precision & Recall & F$1$ \\ \hline

 S.B. &  $0.66$ & $0.97$ & $0.79$ \\ \hline
 RANSAC \cite{ransac} & $0.16$ &  $0.92$ & $$ $0.27$ \\ \hline

\end{tabular}
\end{table}
\section{Practical Applications}
\subsubsection{\textbf{Warehouse Automation}}
A large number of novel items arrive in the warehouses on a daily basis. Dealing with such items becomes a major challenge because not every item can be included in the dataset of the learning based perception algorithms. In such cases, the binary surface classification using shape back-projection can be utilized in order to segregate items in the cluttered containers on the basis of their 3D shape and the segregated items then can be sent to different destinations for further processing.
%
%
%
%
%
%
%
\subsubsection{\textbf{Suction Grasp Location Estimation}}
\begin{figure}[t]
\centering
\includegraphics[width=51ex, height=19ex]{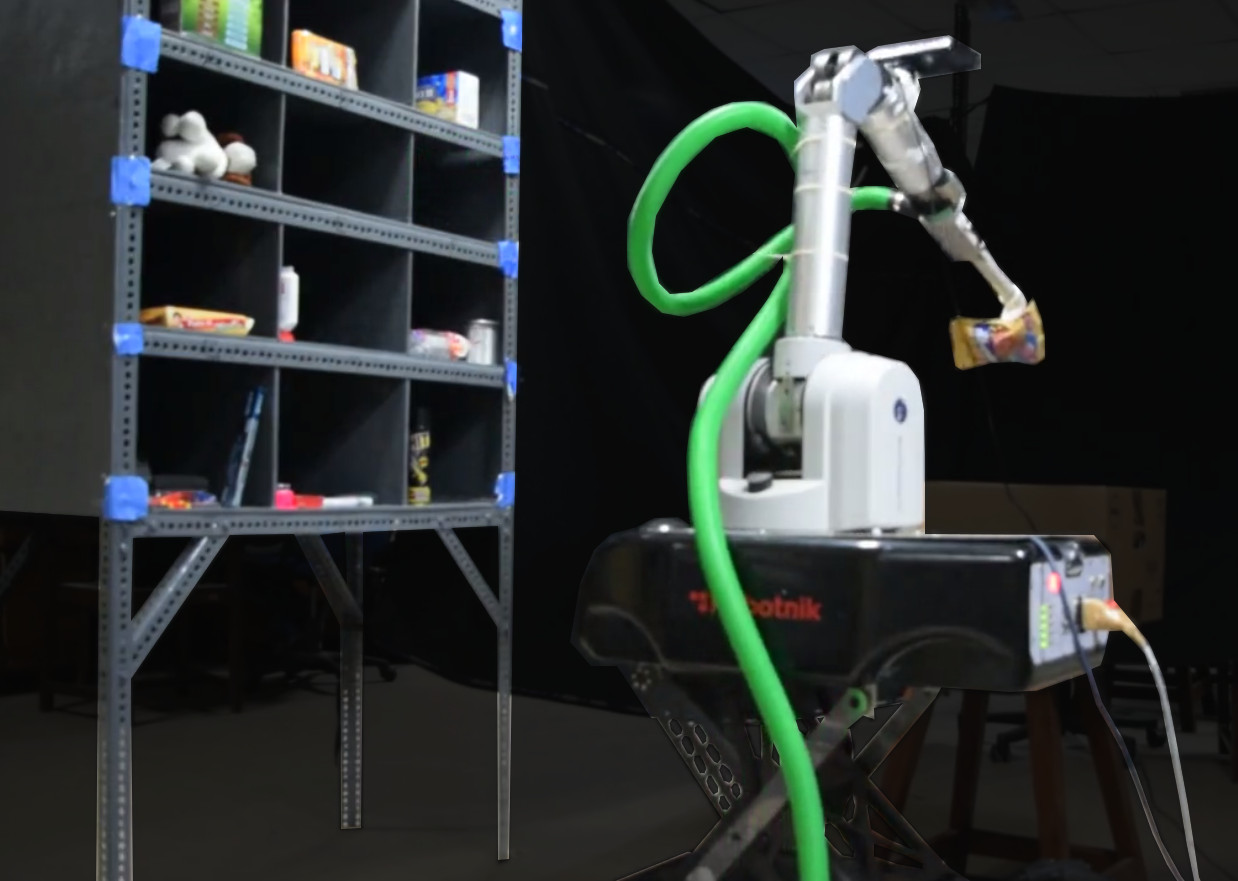}
\caption{Our robotic setup performing an autonomous picking operation on the grasp location provided by shape back-projection.}
\label{fig_suction_grasp}
\end{figure}

Through our participation in the Amazon challenges APC'$16$ and ARC'$17$, we have realized that the centroid based approach \cite{centroidgrasp} is not suitable for suction based grasping in the presence of partial occlusions, and especially when a smaller object lies on top of a larger object. As a solution, shape-histogram of a planar or curved surface (depending on the target object shape) is back-projected onto the point cloud of a target object. Then, an adaptive mean-shift operation is performed on the back-projection likelihood and its convergence point is chosen as the suction grasp location (Fig. \ref{fig_suction_grasp}). This strategy is quite fast, effective and also eliminates a need of time consuming 6D pose estimation such as Super4PCS \cite{super4pcs}.
\subsubsection{\textbf{Automated data labeling for 3D-CNNs}}
In order to train 3D-CNNs for the task of surface classification, a huge amount of data is required. Manual labeling of 3D point clouds is quite exhaustive and challenging as compared to images and requires specialized softwares. Hence, shape back-projection can be deployed to segment specific kind of surfaces from a 3D scene and its output can be used to generate ground truths. For example, the outputs of algorithm in Fig.\ref{fig_shape_backproj_examples} can be used to generate dataset to train a 3D CNN for purpose of surface segmentation and to detect edges.
\section{Conclusion}
In this work, we have presented a novel algorithm of shape back-projection in 3D scenes. Its inspiration lies in the working of color back-projection which obtains color similarity between two images by analyzing their color histograms. Here, we have developed a novel shape histogram, by means of which, a probabilistic measure of similarity between two 3D point clouds can be obtained. The utility of the shape back-projection ranges from warehouse automation to automated labeled data generation for 3D-CNNs. The existing feature based 3D object detection methods can also be benefitted by extracting salient points using shape back-projection. The algorithm is a robust and a computationally efficient alternative to the SOA algorithms for the above purposes. Therefore, it can be easily deployed on computationally limited platforms (UAVs) for complex tasks of object manipulation.

\bibliographystyle{ieeetr}
\bibliography{bibfile}


\end{document}